\documentclass[runningheads]{llncs}

\usepackage[mobile]{eccv}

\usepackage{eccvabbrv}

\usepackage{graphicx}
\usepackage{booktabs}

\usepackage[accsupp]{axessibility}  
\usepackage{multirow}

\usepackage{hyperref}

\usepackage{orcidlink}
\usepackage{pifont}
\definecolor{darkred}{rgb}{0.7, 0.0, 0.0}
\definecolor{darkgreen}{rgb}{0.0, 0.37, 0.14}
\definecolor{darkblue}{rgb}{0.10, 0.17, 0.8}

\begin{document}

\title{AutoDIR: Automatic All-in-One Image Restoration with Latent Diffusion} 

\titlerunning{AutoDIR}

\makeatletter
\newcommand{\printfnsymbol}[1]{%
  \textsuperscript{\@fnsymbol{#1}}%
}
\makeatother
\author{\;\; Yitong Jiang\printfnsymbol{1} \thanks{ \printfnsymbol{1} Equal contribution.}
 \;\;  Zhaoyang Zhang\printfnsymbol{1}  \;\; Tianfan Xue  \;\;  Jinwei Gu \;\; }
\renewcommand\footnotemark{}

\institute{The Chinese University of Hong Kong \\
{\tt\small \{ytjiang@link, zhaoyangzhang@link, tfxue@ie, jwgu@cse\}.cuhk.edu.hk }}


\maketitle

\begin{abstract}
%

%
%

We present AutoDIR, an innovative all-in-one image restoration system incorporating latent diffusion.
AutoDIR excels in its ability to automatically identify and restore images suffering from a range of unknown degradations. 
AutoDIR offers intuitive open-vocabulary image editing, empowering users to customize and enhance images according to their preferences.
%
AutoDIR consists of two key stages: a Blind Image Quality Assessment (BIQA) stage based on a semantic-agnostic vision-language model which automatically detects unknown image degradations for input images, an All-in-One Image Restoration (AIR) stage utilizes structural-corrected latent diffusion which handles multiple types of image degradations. 
Extensive experimental evaluation demonstrates that AutoDIR outperforms state-of-the-art approaches for a wider range of image restoration tasks. The design of AutoDIR also enables flexible user control (via text prompt) and generalization to new tasks as a foundation model of image restoration.  
%
%
\end{abstract}

\section{Introduction}
\label{sec:intro}
Restoring scene details and improving image quality is often the very first step of any computer vision system, which significantly determines the overall system performance. Depending on the operating environments and the signal processing pipelines, images captured by real-world computer vision systems often undergo multiple unknown degradations, such as noise, resolution loss, motion blur, defocus, color imbalance, chromatic aberration, haze, flare, distortion, etc.  
%
Previous image restoration methods often address these basic image degradations separately, including deblurring~\cite{singletaskji2022xydeblur,whang2022deblurring}, denoising~\cite{singletaskquan2020self2self,singletaskzhang2023real} deraining~\cite{singletaskjiang2020multi,singletaskren2019progressive}, super-resolution~\cite{singletaskchen2022femasr,singletaskwang2023exploiting}, low-light enhancement \cite{singletaskyao2023csnorm,singletaskzhang2021star}, deraindrop~\cite{singletaskqian2018attentive}, and dehaze~\cite{singletaskqin2020ffa,singletasksong2023vision}, by using specific single-task models. 
While these single-task approaches yield promising results within their respective tasks, they encounter difficulties when applied to complex real-world situations that involve multiple unknown degradations or require multiple steps of enhancement.

\begin{figure}[!t]
    \includegraphics[width=\textwidth]{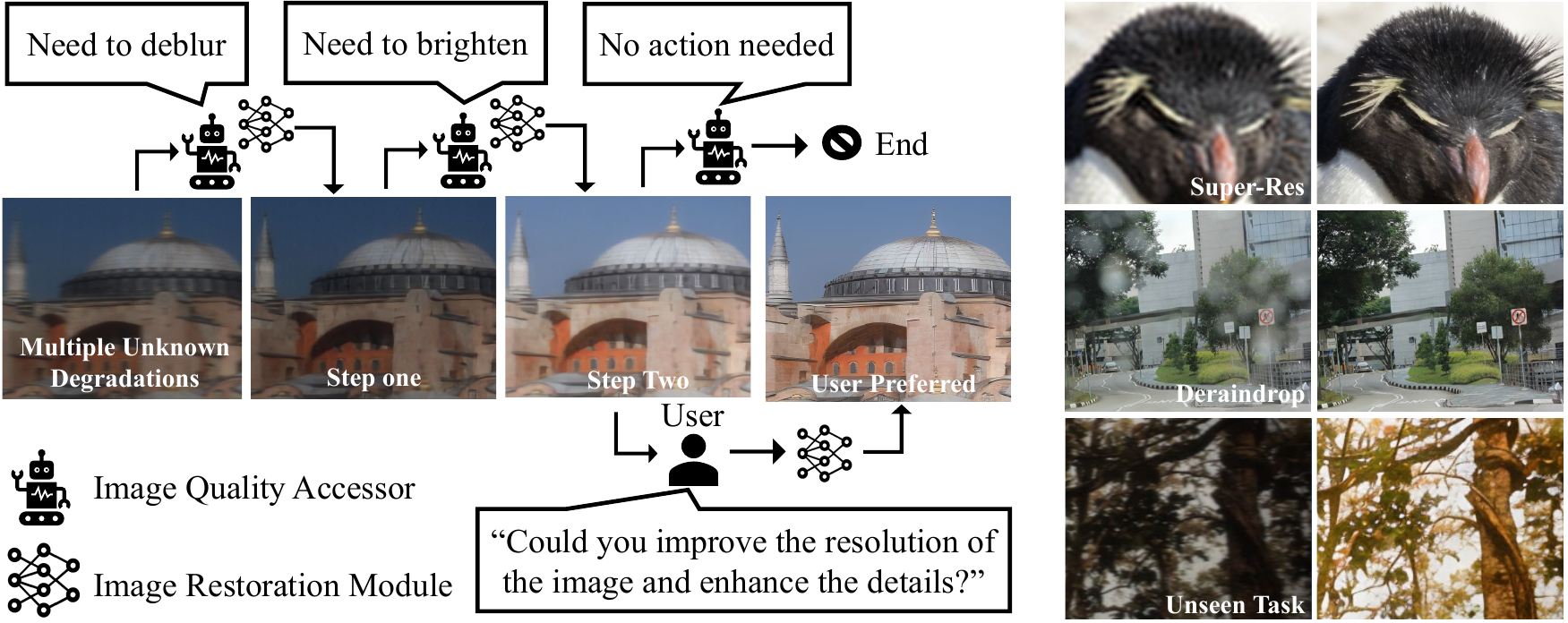}
    \captionof{figure}{
    We propose AutoDIR, an automatic all-in-one model for image restoration capable of handling multiple types of image degradations. 
    \textbf{Left}:  For images with multiple unknown degradations, AutoDIR automatically decomposes the task into multiple steps and supports user interaction via an intuitive text prompt. \textbf{Right}:  AutoDIR effectively restores clean images from different degradations and can handle images with unknown degradations in unseen tasks.
    \textbf{(Please zoom in for details)}}
    \label{fig:representative}
    \vspace{-5mm}
\end{figure}

In this paper, we explore a generalist model capable of handling multiple unknown degradations for a single image. To achieve this goal, the corresponding model should possess the following abilities: (1) decomposing and distinguishing unknown degradations, (2) restoring various degradations in a task-agnostic framework, and (3) ideally, allowing users to freely tune restoration results according to their visual preferences. Numerous related efforts have been made to solve similar problems, but none of them can simultaneously satisfy all three points.

Recent works \cite{cheon2021perceptual,yao2019attention,8576582} have explored assessing unknown degradations by training an image classifier and using this information to explicitly select a specific image restoration model\cite{chen2021pre,liang2021swinir,zamir2022restormer}. However, these methods require training a specific model for each task and primarily focus on a small set of tasks (e.g., 3 tasks). Experiments demonstrate that on large datasets with a wide range of degradations, training an accurate image degradation classifier solely based on image information is challenging (See Tab. \ref{tab:SA}).

Other works attempt to unify multiple image restoration methods \cite{wang2023ddnm, kawar2022denoising} in a single model by utilizing generative model priors. These methods, however, either require known degradation formulations or can only handle images with a single degradation from a predefined small set~\cite{AirNet, potlapalli2023promptir}, thus unable to implicitly distinguish multiple unknown degradations.

In this work, we propose a pipeline named AutoDIR, which satisfies all three aforementioned abilities and can automatically detect and restore images with multiple unknown degradations. AutoDIR consists of two stages: the Semantic-Agnostic Blind Image Quality Assessment (SA-BIQA) stage and the All-in-One Image Restoration (AIR) stage guided by text prompts generated in SA-BIQA.

In the SA-BIQA stage, we enable accurate identification of each degradation in cases of unknown artifacts in an open-vocabulary manner. This is achieved through our proposed Semantic-Agnostic CLIP (SA-CLIP) model, which employs a Semantic-Agnostic regularization term to transfer the original Semantic-Recognitive CLIP to a semantic-agnostic form, focusing on image structural quality rather than semantic content. For detailed information, please refer to Sec.~\ref{sec:biqa}.
%
%
Additionally, we can utilize the text embeddings generated by the SA-BIQA stage as instructions to guide the further restoration model. This approach can not only enable effective restoration but also allows for flexible user control and editing at runtime by providing open-vocabulary instructions, as shown in Fig.~\ref{fig:representative}.

The AIR stage is to handle   degradations using a multitasking image restoration model jointly trained on a wide range of tasks. Given the diverse nature of different tasks (e.g., some tasks like super-resolution need to hallucinate textures, but other tasks like low-light enhancement need to preserve everything except brightness), we propose a hybrid approach that maximizes the hallucinating ability of diffusion models while preserving the image structural consistency by involving extra structural inductive bias.

To evaluate the effectiveness and generalization ability of AutoDIR, we have conducted a comprehensive set of experiments encompassing seven image restoration tasks, including denoising, motion deblurring, low-light enhancement, dehazing, deraining, deraindrop, and super-resolution. Experimental results show that AutoDIR consistently outperforms state-of-the-art methods. AutoDIR is also evaluated for the restoration of images captured by under-display cameras and underwater cameras, which are examples of imaging systems with multiple unknown degradations. Please refer to Sec.~\ref{sec:experiments_new} for details.

\section{Related Work}
\label{sec:related}

\paragraph{\textbf{Unified All-in-One Image Restoration}}
Previous approaches for unified all-in-one image restoration can be categorized into two main groups: unsupervised generative prior-based methods~\cite{wang2023ddnm,pan2021exploiting,yang2021gan,kawar2022denoising,fei2023generative, priorhussein2020image, priorbau2019semantic, priorchan2021glean, priormenon2020pulse} and end-to-end supervised learning-based methods \cite{valanarasu2022transweather,li2020all,zhang2023ingredient,park2023all,AirNet, chen2022learning}. 
Generative priors for image restoration have been widely exploited for image restoration, in the form of generative adversarial networks (GANs) inversion and filters~\cite{priorhussein2020image,priorbau2019semantic,priorchan2021glean,priormenon2020pulse,gu2020image,pan2021exploiting, abu2022adir}, Vector Quantization codebooks~\cite{van2017neural, zhou2022towards,singletaskchen2022femasr,liu2023learning}, latent diffusion models~\cite{wang2023ddnm,kawar2022denoising,fei2023generative, abu2022adir}. Nevertheless, these methods rely on specific known image degradation functions, which constrain the types of degradations they can handle.
%
%
%
End-to-end supervised learning-based methods typically learn image embeddings extracted by an auxiliary degradation predictor to guide the image restoration model \cite{valanarasu2022transweather,zhang2023ingredient,park2023all, AirNet}. For example, Valanarasu et al. \cite{valanarasu2022transweather} leverage a transformer encoder to capture hierarchical features of haze, while Li et al. \cite{AirNet} employ a degradation classifier trained using contrastive learning. Park et al. \cite{park2023all} design a degradation classifier for multiple types of degradation to select appropriate Adaptive Discriminant filters, altering network parameters based on specific degradations.
However, these methods are constrained to handle limited image degradation categories and cannot restore images with multiple unknown degradations.
In contrast, with an SA-BIQA stage, the proposed AutoDIR is able to unify complex degradations without relying on assumptions about the specific degradation functions.

\paragraph{\textbf{Text-based Image Synthesis and Editing}}
Early works~\cite{tao2022df, xu2018attngan, ye2021improving, zhang2021cross, zhu2019dm} on text-to-image synthesis and editing primarily relied on generative adversarial networks (GANs)~\cite{goodfellow2014generative}. 
Recently, developments in diffusion models~\cite{ho2020denoising, rombach2022high} and language-vision models~\cite{radford2021learning} have led to significant progress in text-based image synthesis and editing~\cite{nichol2021glide, ramesh2022hierarchical, saharia2022photorealistic, chefer2023attend}.
The emergence of diffusion-based methods has offered new avenues for text-to-image editing, which can be broadly categorized into data-driven \cite{brooks2023instructpix2pix} and no-extra-data-required approaches \cite{kawar2023imagic, zhang2023sine, hertz2022prompt}.
In the data-driven category, Brooks et al.~\cite{brooks2023instructpix2pix} employ a stable diffusion model fine-tuned with a large dataset of text prompt and image pairs. However, it is primarily designed for semantic image editing and struggles to deliver satisfactory results in tasks related to image quality enhancement. 
On the other hand, some methods focus on image editing without extra training data. For instance,  recent works interpolate the text embedding~\cite{kawar2023imagic} or score estimate~\cite{zhang2023sine} of the input and desired images. However, it requires individual finetuning of the diffusion model for each input image, leading to time-consuming operations. 
Hertz et al.~\cite{hertz2022prompt} improve editing efficiency by directly manipulating cross-attention maps without the need for per-image finetuning. However, it requires interior maps in the reversing process and is thus not applicable to real image editing.
Unlike the above methods, AutoDIR supports real image enhancement without per-image finetuning. 
\begin{figure}[h!]
    \centering
    \includegraphics[width=\columnwidth]{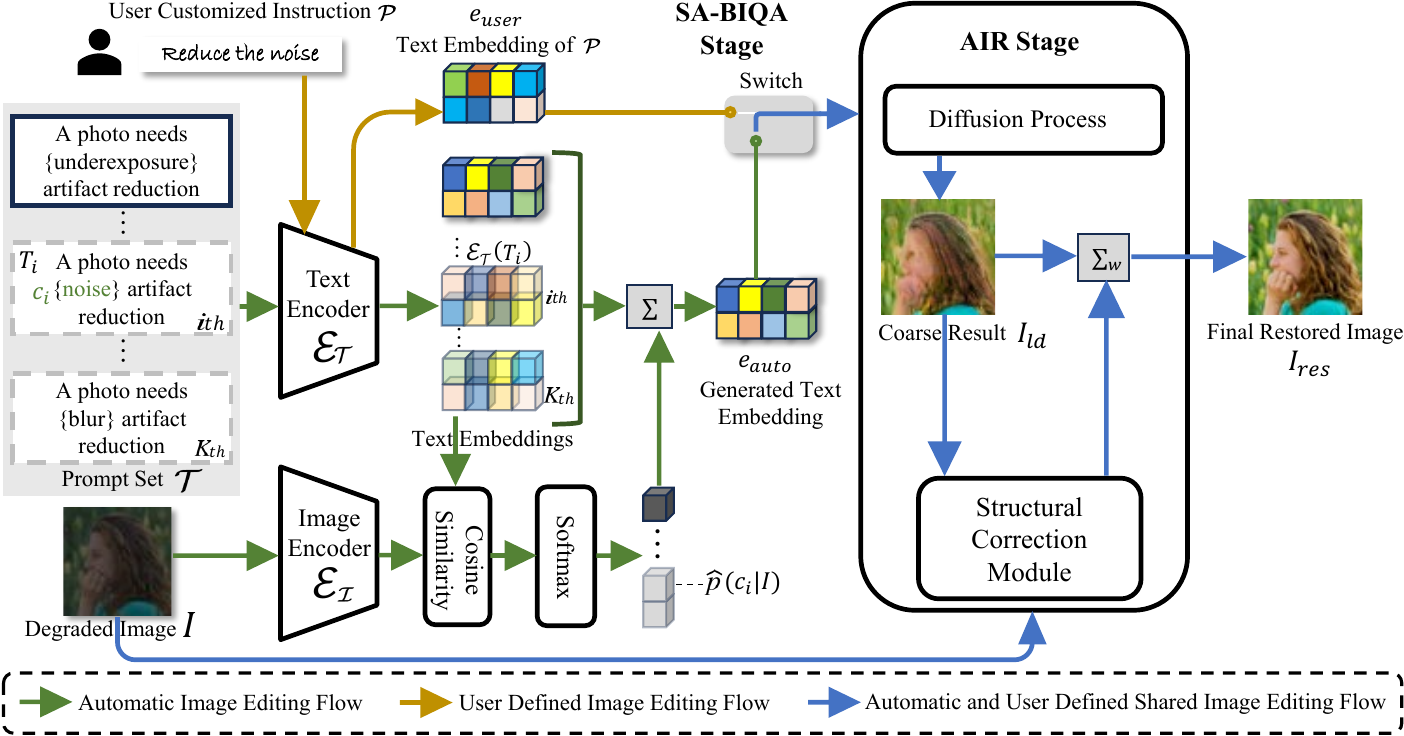}
    \caption{Diagram of the proposed All-in-One Image Restoration with Latent Diffusion (AutoDIR). Refer to Sec.~\ref{sec:method} for more details.}
    \label{fig:framework}
        \vspace*{-6mm}    
\end{figure}

\section{Our Method}
\label{sec:method}
Figure~\ref{fig:framework} shows the overall flowchart of the proposed AutoDIR (Automatic All-in-One Image Restoration with Latent Diffusion), a unified model that can automatically detect and address multiple unknown degradations in images. AutoDIR comprises two main stages:
\begin{itemize}
    \item Semantic-Agnostic Blind Image Quality Assessment (SA-BIQA): This stage automatically identifies the predominant degradations present in the input image, such as noise, blur, haze, and generates a corresponding text prompt, denoted as $e_{auto}$, which is subsequently used in the image restoration process.
    \item  All-in-one Image Restoration (AIR): This stage utilizes a  Structural-Correction Latent Diffusion Model(SC-LDM) to produce a restored image $I_{res}$, guided by the text embedding $e_{auto}$ from SA-BIQA or a user-defined open-vocabulary instruction $e_{user}$. 
\end{itemize}

\subsection{Semantic-Agnostic Blind Image Quality Assessment (SA-BIQA)}
\label{sec:biqa}
As demonstrated by previous works~\cite{cheon2021perceptual,yao2019attention,8576582}, a common approach to assess image degradation is to train an image classifier specifically for that purpose. However, this naive approach faces challenges when dealing with large datasets containing a wide range of artifacts. Even with the use of a heavy ViT encoder, the accuracy of classifying image degradation solely based on image information is limited to 77.65\%, as shown in Tab.~\ref{tab:SA}.

To address this limitation, we propose leveraging human language knowledge to enhance the detection of image degradation. We introduce a Semantic-Agnostic CLIP (SA-CLIP) model as our backbone for Blind Image Quality Assessment (BIQA). SA-CLIP builds upon the CLIP model, which establishes a connection between human language knowledge and image quality.
However, we observed that directly applying CLIP or naively fine-tuning it for BIQA tasks does not yield reliable results, as shown in Tab.~\ref{tab:SA}. We delved into this problem and identified the reason behind this issue. The pre-trained CLIP model is primarily trained for visual recognition tasks, which prioritize semantic information rather than image quality. Consequently, it delivers low accuracy in BIQA tasks. For instance, the model may struggle to differentiate between a low-light dog image and a noisy dog image because it focuses more on the "dog" aspect rather than the presence of noise or illumination.

To overcome this, we tackle the problem in Two steps:
(i) We construct a new image quality assessment task for fine-tuning CLIP.
(ii) We propose a new regularization term for semantic-agnostic and image-quality-aware training to derive SA-CLIP model.

As illustrated in Fig.~\ref{fig:framework}, 
let $\mathcal{C}$ denotes the set of image distortions, i.e. $ \mathcal{C} = \{c_{1}, c_{2}, ..., c_{K-1}, c_{K}\}$, where $c_{i}$ is an image distortion type, e.g. ``noise'' and $K-1$ is the number of distortions we consider. We also add a special type $c_{K}$ =``no'' to indicate the end of the automatic multiple-step restoration.
 The textural description prompt set is $\mathcal{T} = \{T $ $| T =$ ``A photo needs $\{c_i\}$ artifact reduction.'', $c \in \mathcal{C}\}$. 
Given a corrupted image $I$ which undergoes several unknown artifacts, our Semantic-Agnostic CLIP aims to identify the dominant degradation of $I$ and extract the corresponding text embedding. 
SA-CLIP consists of an image encoder $\mathcal{E_I}$
and a text encoder $\mathcal{E_T}$. We first obtain the image embedding $\mathcal{E_I}( I) \in \mathbb{R}^d $ and the text embedding $\mathcal{E_T}(T) \in \mathbb{R}^{K \times d}$, then we compute the cosine similarity between image embedding $\mathcal{E_I}( I)$ and each prompt embedding $\mathcal{E_T}(T_i) \in \mathbb{R}^d $ by:
\begin{equation}\label{eq:logits}
     \textrm{logit}(c_i | I) =  \frac {\mathcal{E_I}( I) \cdot \mathcal{E_T}(T_i)}{ \|\mathcal{E_I}( I)\|_2 \| \mathcal{E_T}( T_i)\|_2 },
\end{equation}
where $T_i$ is the $i$-{th} element of $ \mathcal{T}$. Next, we calculate the probability $\hat{p}(c_i | I)$ and obtain the output text embedding of BIQA ${e}_{auto}$ by:  
\begin{equation}\label{eq:probability}
\hat{p}(c_i | I) = \frac{\exp(\textrm{logit}(c_i | I))}{\sum_{i=1}^{K} \exp(\textrm{logit}(c_i | I))},
\end{equation}
\begin{equation}\label{eq:textout}
{e}_{auto} = \sum_{i=1}^{K} \hat{p}(c_i | I)  \mathcal{E_T}( T_i).
\end{equation}
\paragraph{\textbf{Naive fine-tuning on image quality assessment}}
During the optimization of CLIP model, we freeze the parameters of the text encoder $\mathcal{E_T}$ and finetune the image encoder $\mathcal{E_I}$ using the multi-class fidelity loss \cite{tsai2007frank}. The fidelity loss can be denoted as:
\begin{equation}\label{eq:cliploss}
L_{FID} = 1 - \sum_{i=1}^{K} \sqrt{y(c_i| I) \hat{p}(c_i| I)},
\end{equation}
where $y(c_i| I)$ is a binary variable that equals 1 if $c_i$ is the dominant degradation in image $I$, and 0 otherwise. 
\paragraph{\textbf{Semantic-Agnostic Constraint fine-tuning on image quality assessment}}



\begin{figure}[t]
   
    \includegraphics[width=\textwidth]{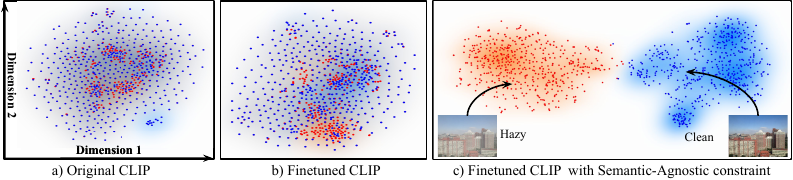}
    \caption{t-SNE visualization of image embeddings  $\mathcal{E_I}( I)$ of CLIP of Blind Image Quality Assessment (BIQA) on SOTs dehazing dataset \cite{li2018benchmarking}. Image embeddings of foggy images and their ground-truth clean images are extracted by a) original CLIP. b) finetuned CLIP. c) SA-CLIP finetuned with semantic-agnostic constraint. This illustrates that semantic-agnostic constraint can separate the embeddings of the degraded images from the clean images, while original CLIP and finetuned CLIP features cannot.}
    \label{fig:clip_ab}
    \vspace{-5mm}
\end{figure}



Since the original CLIP model is pre-trained on tasks such as image classification, its corresponding $\mathcal{E_I}(I)$ encoders tend to encode images based on their semantic information (\eg, cat or dog) rather than their image quality (\eg, noisy or clean). When we fine-tune the CLIP model for generating texts for BIQA according to image quality, this becomes a significant limitation. 
As shown in Fig.~\ref{fig:clip_ab} a) and b), the image embeddings extracted by the original CLIP and the finetuned CLIP on foggy, as well as their corresponding ground-truth clean images, cannot be separated, indicating a focus on semantic information rather than image quality differences.

To address this issue, we propose a novel approach called the semantic-agnostic constraint loss ($L_{SA}$) to regulate the fine-tuning process and prevent the model from relying solely on semantic information rather than image quality. 
The semantic-agnostic loss ($L_{SA}$) applies a penalty when the CLIP model suggests that the artifact $c_i$ is present in the ground-truth clean image  $I_{gt}$ which corresponds to the degraded image $I$. This penalty forces the CLIP model to distinguish between $I_{gt}$ and $I$ based on image quality, encouraging the CLIP image encoder ($\mathcal{E_I}(I)$) to focus on extracting image quality information rather than semantic information.
This constraint loss can be derived using the following equation:
\begin{equation}\label{eq:cliploss}
L_{SA} = \sum_{i=1}^{K} \sqrt{y(c_i|I)\hat{p}(c_i| I_{gt})},
\end{equation}
where $y(c_i|I)$ is a binary variable that is equal to 1 if $c_i$ represents the dominant degradation in image $I$, and 0 otherwise, and $\hat{p}(c_i|I_{gt})$ denotes the probability that the dominant degradation in $I_{gt}$ is represented by $c_i$.
%

We incorporate the semantic-agnostic constraint $L_{SA}$ with the fidelity loss $L_{FID}$, resulting in the total loss $L_{BIQA}$ for SA-CLIP model:
\begin{equation}\label{eq:cliploss_total}
L_{BIQA} =  L_{FID} + \lambda L_{SA},
\end{equation}
where we set $\lambda=1$ in our experiments.
As shown in Fig.~\ref{fig:clip_ab} c), CLIP fine-tuned with a semantic-agnostic constraint can separate the embeddings of the degraded images from the clean images as expected. 



\setlength{\tabcolsep}{2pt}
\begin{table}[ht!]
		\setlength{\abovecaptionskip}{0pt} \setlength{\belowcaptionskip}{0pt}
\caption{\small Average accuracy of image degradation detection on seven degradation tasks and clean image. For details, please refer to the supplementary.}
\label{tab:SA}
\centering
\footnotesize
\def\arraystretch{1.4}
\scalebox{0.83}{
\begin{tabular}{c|cccc}\toprule

  Methods    & ViT-classifier & Original CLIP & Fine-tuned CLIP & \textbf{ SA-CLIP (ours) }\\
\midrule
 Average Accuracy  & $77.65\%$  &$61.96\%$  & $83.31\%$ & $97.94\%$ \\
  \bottomrule
\end{tabular}}
        \vspace{-10mm}
\end{table}

\subsection{Effectiveness of Semantic-Agnostic CLIP}
We compare our SA-CLIP with a ViT-based~\cite{dosovitskiy2020image} classifier, original CLIP, and naive fine-tuned CLIP to demonstrate the effectiveness of the Semantic-Agnostic training scheme.
Tab.~\ref{tab:SA} presents the effectiveness of our Semantic-Agnostic training scheme. The results demonstrate that our SA-CLIP, which incorporates human language information, achieves higher accuracy compared to the ViT-based classifier which considers only image information. Moreover, when CLIP is fine-tuned with Semantic-Agnostic regularization, the model focuses more on image quality information rather than semantic information, resulting in superior performance compared to the original pre-trained CLIP and CLIP fine-tuned without Semantic-Agnostic regularization. For a visual demonstration of the effectiveness of the Semantic-Agnostic constraint, please refer to the supplementary materials.
\begin{figure}[t]
    \includegraphics[width=\textwidth]{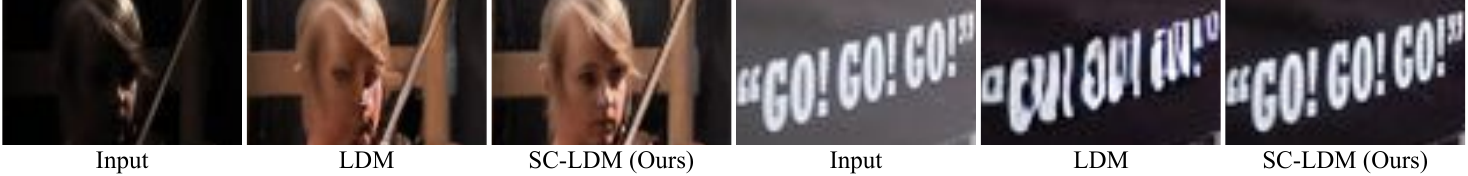}
    \caption{Structural-Correction Latent Diffusion Model (SC-LDM) can maintain complex structures of the original images, while Latent Diffusion Model (LDM) fails to.}
    \label{fig:SCM}
\end{figure}

\subsection{All-in-one Image Restoration (AIR)}
\label{sec:all-in-one}
\paragraph{\textbf{Naively applying LDM on multi-task image restoration}}
The All-in-One Image Restoration (AIR) stage aims to handle multiple degradations in a single shared framework. Recent advancements in diffusion-based generative models \cite{brooks2023instructpix2pix, rombach2022high} have demonstrated their remarkable ability to generate diverse images, making them suitable for multi-task image restoration. Previous studies\cite{wang2023stableSR, chen2022femasr} have shown that generative models possess a remarkable ability to generate missing or distorted details, especially for tasks that require hallucination, such as super-resolution.
Based on these insights, we conduct the AIR stage based on the Latent Diffusion Model (LDM)~\cite{rombach2022high}.
LDM incorporates both text and image embedding conditions to restore images $I_{sd}$ using generative priors. The text embedding condition $e = \{ e_{auto}, e_{user} \}$ aims to separate different types of image degradations, while the latent image embedding condition $z_I = \mathcal{E}_{ldm}(I)$ from the image encoder $\mathcal{E}_{ldm}$ of the LDM provides structural information.

However, although LDM-based generative models can provide a foundation for multi-task image restoration, they have limitations in reconstructing images with complex and small structures due to the compression-reconstruction process with Variational Autoencoder (VAE)~\cite{kingma2013auto}, as discussed in ~\cite{chen2024textdiffuser, zhang2023brush, huang2023collaborative}. ~\cite{huang2023collaborative} attempted to reduce distortion caused by the compression-reconstruction process by retraining Variational Autoencoder (VAE) networks on category-specific images (e.g., human faces) to learn specialized probability distributions. However, this approach is not suitable for image restoration tasks due to the diverse image content. To address these limitations, we introduce a lightweight plug-in structural-correction module to the LDM, enhancing its ability to handle complex and small structures during image restoration.


%

\begin{figure}[t!]
    \centering
\begin{minipage}[a]{0.48\textwidth}
\vspace{-7.1cm}
    \includegraphics[width=\textwidth]{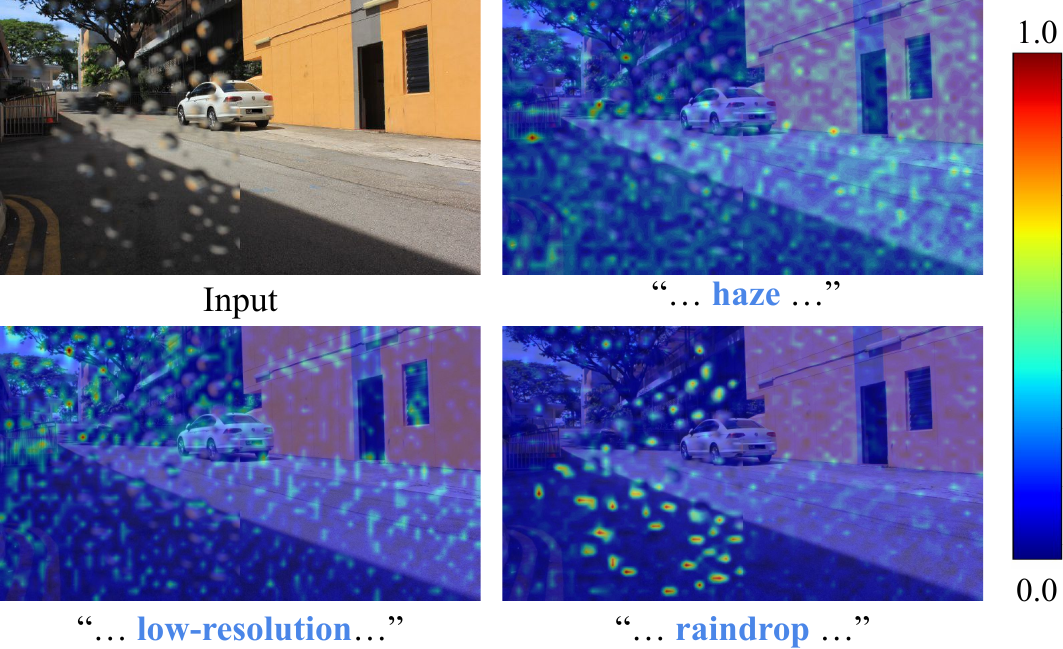}
    \captionof{figure}{Cross-attention maps of text-conditioned diffusion image restoration. The top-left shows an input image with raindrops on the left half. The remaining plots show the cross-attention masks for the keywords ``haze'', ``low-resolution'', and ``raindrop'' in the text prompt used for restoration. While the cross-attention maps for ``haze'' and ``low-resolution'' are more uniformly distributed over the entire image, the map for ``raindrop'' correctly focuses on the actual image artifacts, as expected.} 
    \label{fig:cross_att}
    \end{minipage}
\hfill
\begin{minipage}[b]{0.48\textwidth}
\vspace{0cm}
    \includegraphics[width=\textwidth]{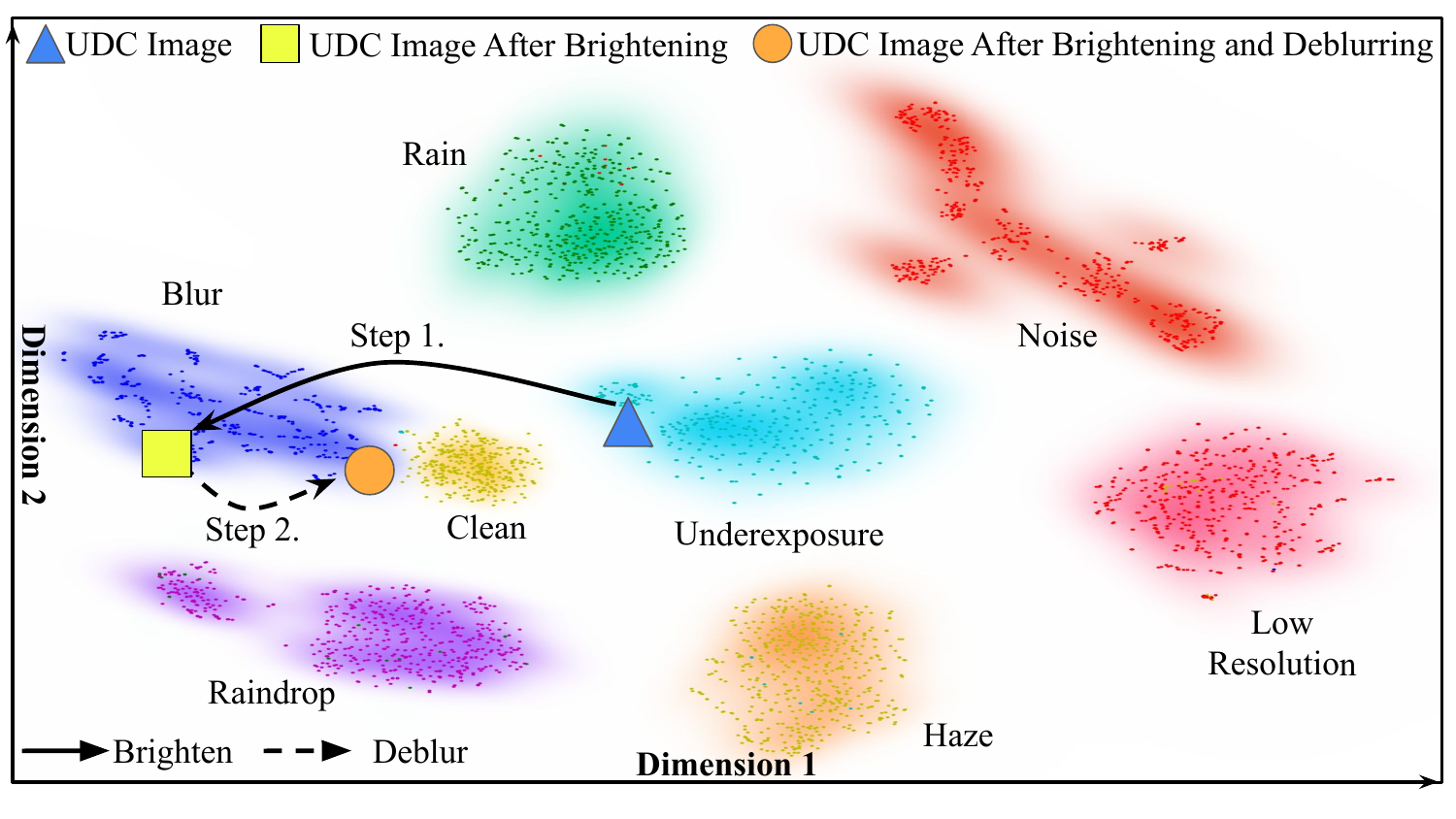}
    \captionof{figure}{t-SNE visualization of image embeddings $\mathcal{E_I}( I)$ with seven types of degradations, which illustrates the space of common image degradations. Images captured by Under-Display Cameras (UDC)~\cite{zhou2021image} suffer from both blur and underexposure. AutoDIR automatically decides (via SA-BIQA) that the first step is to improve ``underexposure'', and the second step is to remove ``blur'', moving the input images towards the region of clean images.} 
    \label{fig:tsne_move}
\end{minipage}
\end{figure}

\paragraph{\textbf{Structural-Correction Latent Diffusion Model (SC-LDM)}}
While the LDM-based generative models can provide a foundation for multi-tasking image restoration, it's widely noticed that they may fail to maintain original image structures such as human faces and texts~\cite{chen2024textdiffuser, zhang2023brush, huang2023collaborative}, as shown in Fig.~\ref{fig:SCM}. 
To address the structure distortion, we employ an efficient structural-correction module (SCM) denoted as $\mathcal{F}$. The purpose of SCM is to extract the contextual information $\mathcal{R}$ from the original image and incorporate it with the intermediate image restoration result $I_{sd}$ in a residual manner. This is achieved through the following equation:
\begin{equation}\label{eq:CBRMR}
I_{res} = I_{sd} + w \cdot \mathcal{F}( [ I_{sd}, I] ),
\end{equation}
where $[~]$ denotes concatenation, and $w$ is an adjustable coefficient that ranges between 0 and 1. The value of $w$ determines the extent to which contextual information is utilized to recover the final result. A larger value of $w$ emphasizes the use of contextual information, which is beneficial for tasks that require structure consistency, such as low light enhancement. Conversely, a smaller value of $w$ is often employed to maintain the generation capability of latent diffusion model for tasks like super-resolution. By integrating SCM, AutoDIR effectively restores the distorted context of the original image, seamlessly incorporating the enhancements made during the editing phase as shown in Fig.~\ref{fig:SCM}. 

During training, we fine-tune the UNet backbone $\epsilon_{\theta}(e, [z_t, z_I], t)$ of the Latent Diffusion Model (LDM) for image restoration tasks. The objective function is defined as:
\begin{equation}\label{eq:OCLDM}
L_{LD} =  \mathbb{E}_{\mathcal{E}_{ldm}(x), c_{I},e,\epsilon ,t} [\| \epsilon - \epsilon_{\theta}(e, [z_t, z_I] , t)\|_2^2].
\end{equation}
For the Structural-Correction Latent Diffusion Model (SC-LDM), instead of generating the edited latent $\hat{z}_t$ with the complete reversing sampling process, which is time-consuming, we utilize the estimated edited latent $\Tilde{z}$, which is calculated by:
\begin{equation}
\Tilde{z} = \frac{z_t}{\sqrt{\bar{\alpha}}} - \frac{\sqrt{1-\bar{\alpha}} \epsilon_{\theta}(e, [z_t, z_I] , t)}{\sqrt{\bar{\alpha}}},
\end{equation}
where $\bar{\alpha}$ represents the noise scheduler introduced in \cite{rombach2022high}. The loss function for Structural-Correction Latent Diffusion Model (SC-LDM) is further defined as:
\begin{equation}
L_{AIR} = \| I_{gt} - \left(\mathcal{F}(\mathcal{D}(\Tilde{z}), I) + \mathcal{D}(\Tilde{z})\right)\|_2^2
\end{equation}

\paragraph{\textbf{Mechanism of handling multi-task image restoration}}
Fig.~\ref{fig:cross_att} shows our experiment of exploring the mechanism of the text conditions disentangling different image restoration tasks
during the reverse diffusion process.
We find that different text conditions yield different cross-attention maps. 
As shown in Fig.\ref{fig:cross_att}, changing the text prompt leads
to significant changes in the cross-attention map. The map closely aligns with the text prompt, which either uniformly distributes attention across the entire image for the “dehazing” prompt, or concentrates on the part with edges or textures for the ``low-resolution'' prompt, or focuses on specific areas such as raindrops for the ``deraindrop'' prompt.
This shows that AutoDIR can guide the diffusion attention toward regions that are more likely to have image artifacts.

\label{sec:experiments}

\section{Experiments}
\label{sec:experiments_new}
To fully validate the feasibility and effectiveness of unifying multiple restoration and enhancement tasks, we compare AutoDIR with both CNN-based ~\cite{chen2022simple, AirNet, potlapalli2023promptir}, and diffusion-based multitasking~\cite{rombach2022high} backbones on two tasks with unknown image degradations as well as seven representative image restoration tasks. Specifically, we highlight the advantages of our generative model-based pipeline over discriminative models for tasks that require hallucinating details such as super-resolution. Please refer to the appendix for implementation details and datasets. The source code and trained models will be released upon publication.

\begin{figure*}[t]
    \centering
    \includegraphics[width=\textwidth]{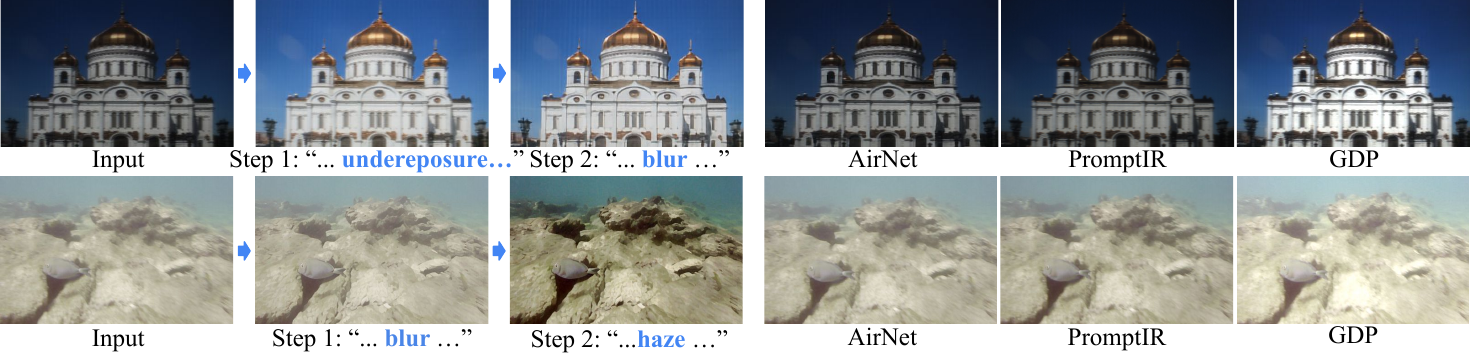}    
    \caption{AutoDIR deals with multiple unknown image degradations. Iterative image restoration on unseen Under-Display Camera (TOLED) dataset~\cite{zhou2021image} and unseen Enhancing Underwater Visual Perception (EUVP) dataset~\cite{islam2020fast}. Left: AutoDIR decomposes and solves the unseen artifact in several steps. Right: all-in-one AirNet~\cite{AirNet}, PromptIR~\cite{potlapalli2023promptir} and zero-shot method GDP~\cite{fei2023generative} for comparison.}
    \label{fig:iterative}
\end{figure*}

\subsection{Results on Multiple Unknown Degradations }
\label{sec:results}
To show that AutoDIR can handle multiple unknown image degradations, we conducted experiments on two additional datasets: the Under-Display-Camera (TOLED) dataset~\cite{zhou2020image} and the Enhancing Underwater Visual Perception (EUVP) dataset~\cite{islam2020fast}. These two datasets consist of multiple unknown image degradations and are completely excluded for training, allowing us to evaluate the performance of AutoDIR in handling unknown, diverse image distortions.
Fig.~\ref{fig:iterative} demonstrates the iterative process of AutoDIR to handle images from these two datasets. As shown, AutoDIR can automatically decompose the UDC restoration in two steps (``brightening'' and ``deblurring''), and the Underwater restoration in two steps (``deblurring'' and ``dehazing''), enabling the restoration of input images. For comparison, we also show the results of a recent all-in-one AirNet~\cite{AirNet}, PromptIR~\cite{potlapalli2023promptir}, diffusion-based zero-shot image restoration method GDP~\cite{fei2023generative}. Furthermore, as shown in Tab.~\ref{tab:UDC}, AutoDIR outperforms all the other black-box and classifier-based all-in-one methods on quantitative perceptual results. We also conducted a user study on 28 images of TOLED and real-captured images by OPPO Reno6 device. We collected 22 forms and 28×22=616 responses. And AutoDIR receives $\textbf{96.9\%}$ of the votes as the best result, please refer to the appendix for details.

\begin{figure*}[t]
    \centering
    \includegraphics[width=0.99\textwidth]{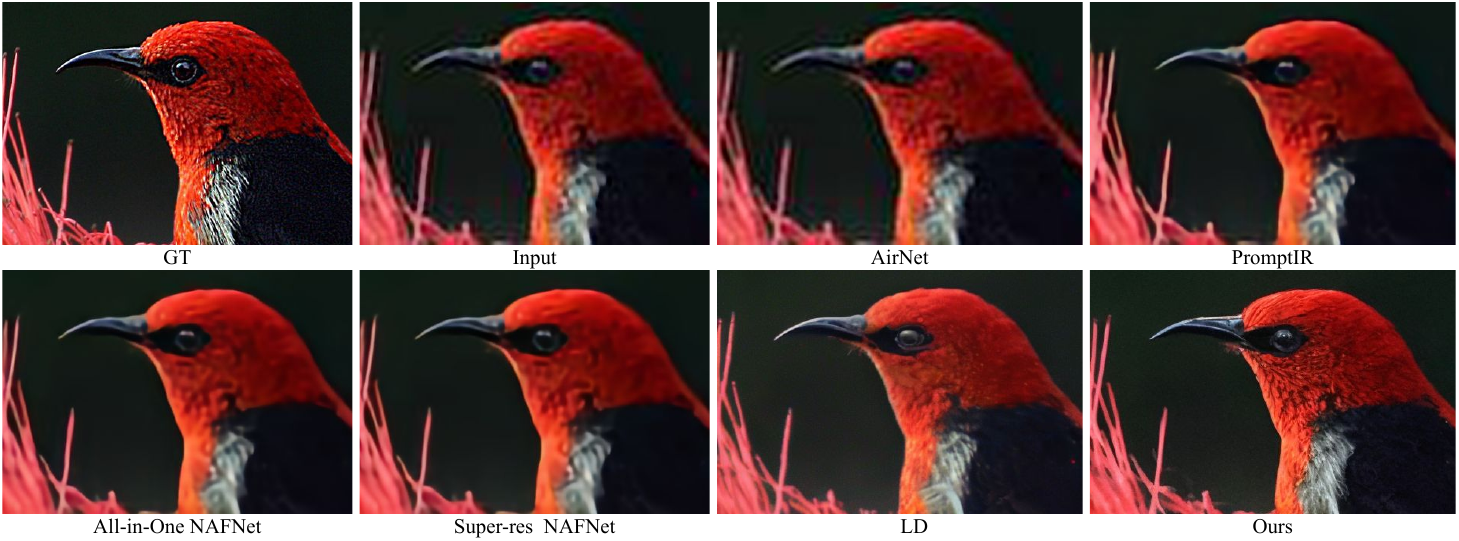}
    \caption{Super-resolution qualitative comparisons with AirNet~\cite{AirNet}, PromptIR~\cite{potlapalli2023promptir}, NAFNet \cite{chen2022simple} trained on all-in-one tasks, NAFNet \cite{chen2022simple} trained on Super-Res, and Latent Diffusion Model (LDM) \cite{rombach2022high}. See the Appendix for more results \textbf{(Zoom in for better view)}.}
    \label{fig:noise}
\end{figure*}

\hspace{-0.65cm}
\begin{minipage}{0.56\textwidth}
    \setlength{\tabcolsep}{2pt}
\vspace{5mm}
\captionof{table}{\small Quantitative comparison on SR}
\label{tab:Comparison2}
\centering
\footnotesize
\def\arraystretch{1.2}
\scalebox{0.65}{
\begin{tabular}{cccccccc}\toprule
  \multirow{2}{*}{\textbf{Method}}   & \multicolumn{5}{c}{\textbf{Super Resolution}}\\

  &\scriptsize MUSIQ $\uparrow$ & \scriptsize CLIP-IQA $\uparrow$ & \scriptsize NIQE $\downarrow$ & \scriptsize NIMA $\uparrow$ & \scriptsize LPIPS  $\downarrow$ & \scriptsize PSNR $\uparrow$ &\scriptsize SSIM $\uparrow$ \\
    \midrule
   NAFNet   & 40.77 & 0.292& 8.45 & 4.45 &\textbf{0.278} &\underline{28.52}  & 
   \underline{0.754} 
\\
   NAFNet-SR   & 45.49 & 0.299& 8.23 & 4.56 &0.332 &{27.95}  & 
   0.726
\\
   AirNet   &20.21 & 0.276 & 11.07 & 4.03 & 0.425 & 28.14 & 0.626 \\
   
   PromptIR & 22.89 & 0.301&10.04 & 4.16 & 0.309 &  \textbf{29.62} & \textbf{0.792} \\

    LD &\underline{50.71} & \underline{0.375} & \underline{6.44} & \underline{4.90}& 0.378 &  {24.11} &0.649\\
  \textbf{Ours} &\textbf{63.02} & \textbf{0.596} &\textbf{4.62} & \textbf{5.25}& \underline{0.282} &  {23.49} & 0.599\\
  \bottomrule
\end{tabular}}
\label{tab:SR}
\end{minipage}
\hfill
\hspace{-2mm}
\begin{minipage}{0.4\textwidth}
\setlength{\tabcolsep}{2pt}
  \vspace{4mm}
\captionof{table}{\small Quantitative comparison on Zero-shot UDC task.}
\label{tab:Comparison2}
\centering
\footnotesize
\def\arraystretch{1.3}
\scalebox{0.6}{
\begin{tabular}{ccccc}\toprule

 \multirow{2}{*}{\textbf{Method}}   & \multicolumn{4}{c}{\textbf{Under-display Camera}}\\ 

  & \scriptsize MUSIQ $\uparrow$ & \scriptsize CLIP-IQA $\uparrow$ & \scriptsize NIQE $\downarrow$ & \scriptsize NIMA $\uparrow$ \\
    \midrule
   NAFNet   & 37.16 & 0.278 & 7.402 & 4.258\\

   AirNet   & 41.20 & 0.265 & 7.681 &4.048 \\
   
   PromptIR  & 45.40 & 0.303 & 8.294 &4.344 \\
   
   LD & 43.41 & 0.288 & 7.983 & 4.371\\
  \textbf{Ours} & \textbf{49.27} & \textbf{0.311} &\textbf{6.392} & \textbf{4.427}\\
  \bottomrule
\end{tabular}}
\label{tab:UDC}
\end{minipage}

\begin{figure*}[t]
    \centering
    \includegraphics[width=1.0\textwidth]{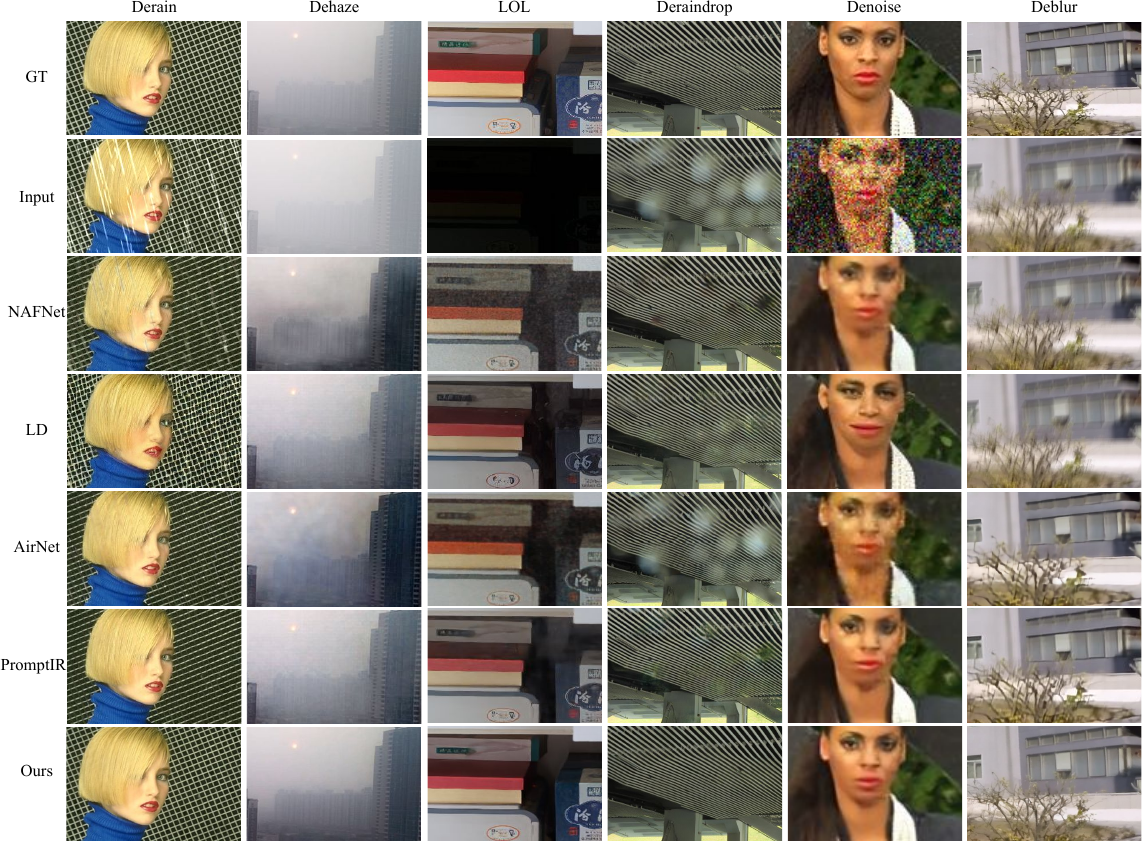}
    \caption{Qualitative comparisons with NAFNet\cite{chen2022simple} and Latent Diffusion Model (LDM) \cite{rombach2022high}, AirNet~\cite{AirNet} and PromptIR~\cite{potlapalli2023promptir} on deraining, dehazing, low light enhancement, and deraindrop, denoise and deblur tasks. See the Appendix for more results. \textbf{(Zoom in for better view)}}

    \label{fig:rain}
    \vspace{-8mm}
\end{figure*}

\subsection{Results on Multi-task Image Restoration}
\label{sec:multi-task}
We compare with state-of-the-art methods: the CNN-based single-task image restoration pipeline NAFNet~\cite{chen2022simple}, CNN-based all-in-one AirNet~\cite{AirNet}, PromptIR ~\cite{potlapalli2023promptir}, and the generative backbone Latent Diffusion~\cite{rombach2022high}. They are all fine-tuned for all seven tasks, employing an identical training configuration to AutoDIR.
%
%
Tab.~\ref{tab:SR} and Tab.~\ref{tab:Comparison1} present quantitative results of AutoDIR and other methods in all seven tasks. 
As demonstrated, AutoDIR achieves competitive performance across all seven tasks. Specifically, to further evaluate the ability of discriminative and generative methods on super-resolution tasks, we include task-specific discriminative NAFNet training on SR. As shown in Tab.~\ref{tab:SR} and Fig.~\ref{fig:noise}, AutoDIR quantitatively and qualitatively outperforms all the all-in-one discriminative, task-specific discriminative methods and the black-box latent diffusion model.  Please be noted that, as discussed in previous works~\cite{yu2024scaling,blau2018perception,jinjin2020pipal}, full-reference metrics like PSNR, SSIM, and LPIPS~\cite{zhang2018perceptual} often do not align with human evaluation on super-resolution tasks. Therefore, following~\cite{yu2024scaling,wang2023stableSR}, we include non-reference metrics MUSIQ~\cite{ke2021musiq}, CLIP-IQA~\cite{wang2022exploring}  NIQE~\cite{mittal2012making} and NIMA~\cite{talebi2018nima} to better evaluate the results.
%
We also show the qualitative comparison results on the other six image restoration tasks in Fig.~\ref{fig:rain}, which demonstrate that AutoDIR can achieve more visually appealing results than the other baselines.
A comprehensive and detailed analysis of the quantitative and qualitative results for each specific task can be found in the appendix.

\setlength{\tabcolsep}{3.5pt}
\begin{table}[t!]
		\setlength{\abovecaptionskip}{0cm}
		\setlength{\belowcaptionskip}{0cm}
\caption{Quantitative comparison on denoising, deraining, dehazing, deraindrop, and low light enhancement tasks. The best results are marked in boldface and the second-best results are underlined.}
\label{tab:Comparison1}
\centering
\footnotesize
\def\arraystretch{1.3}
\scalebox{0.545}{
\begin{tabular}{ccccccccccccccccccc}\toprule
 \multirow{3}{*}{\textbf{Method}} & \multicolumn{3}{c}{\textbf{Denoise}} & \multicolumn{3}{c}{\textbf{Derain} } & \multicolumn{3}{c}{\textbf{Dehaze}}  & \multicolumn{3}{c}{\textbf{Deraindrop}}  & \multicolumn{3}{c}{\textbf{Low light}}  & \multicolumn{3}{c}{\textbf{Deblur}} 
  \\  &\scriptsize SSIM$\uparrow$ & \scriptsize PSNR$\uparrow$ & \scriptsize LPIPS$\downarrow$ & \scriptsize SSIM$\uparrow$ & \scriptsize PSNR$\uparrow$ & \scriptsize LPIPS$\downarrow$ & \scriptsize SSIM$\uparrow$ & \scriptsize PSNR$\uparrow$ & \scriptsize LPIPS$\downarrow$ & \scriptsize SSIM$\uparrow$ & \scriptsize PSNR$\uparrow$ & \scriptsize LPIPS$\downarrow$  & \scriptsize SSIM$\uparrow$ & \scriptsize PSNR$\uparrow$ & \scriptsize LPIPS$\downarrow$  & \scriptsize SSIM$\uparrow$ & \scriptsize PSNR$\uparrow$ & \scriptsize LPIPS$\downarrow$\\
 \cmidrule(l){1-1}  \cmidrule(l){2-4} \cmidrule(l){5-7}  \cmidrule(l){8-10} \cmidrule(l){11-13} \cmidrule(l){14-16} \cmidrule(l){17-19}
   NAFNet  & 0.805 & {28.10} &  0.195 &  0.926 & 30.46 & 0.100 & 0.968   & 27.75 & 0.028  & 0.872 & 25.04 &  0.138 & 0.827 & 21.15 & 0.240 & 0.805 & 26.67 & 0.208\\
  
   LD   & 0.625 & 22.58 & 0.186 & 0.651 & 23.21 & 0.147 & 0.763 & 23.49 &  0.091 & 0.738 & 
 24.84 & 0.104 & 0.770 &  18.97 & 0.159 & 0.695 & 22.53 & 0.241\\
 
  {AirNet} & 0.803 & 29.10 &  0.213 & 0.929 & 30.99 & 0.055 & 0.944 & 26.52 & 0.031 & 0.892 & 27.13 & 0.094 & 0.818 & 21.26 & 0.237 & 0.800 & 26.50 & 0.201\\

     PromptIR  & \underline{0.824} & \textbf{29.89} &  \underline{0.181} &  \underline{0.938} & \underline{33.97} & \underline{0.049} &\underline{0.971}   & \underline{29.13} & \underline{0.022}  & \underline{0.900} &\underline{27.41} & \underline{0.081} & \underline{0.831} & \textbf{22.42} & \underline{0.133} & \underline{0.819} & \underline{26.82} & \underline{0.198} \\
 
 \textbf{Ours} &\textbf{0.832} & \underline{29.68} &\textbf{0.167} & \textbf{0.965} & \textbf{35.09
} &\textbf{0.047} &\textbf{0.973} &\textbf{29.34}  &\textbf{0.020} &\textbf{0.924} &\textbf{30.10} &\textbf{0.063}  & \textbf{0.888} &\underline{22.37} &\textbf{0.129}  & \textbf{0.828} &\textbf{27.07} &\textbf{0.157}\\
  \bottomrule
\end{tabular}}
\vspace{2mm}
\end{table}

\begin{figure}
    \centering
    \includegraphics[width=\textwidth]{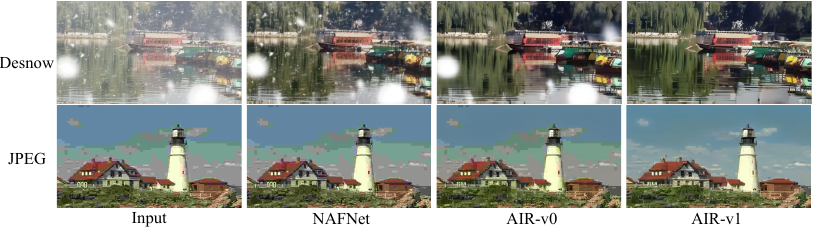}
    \captionof{figure}{Qualitative comparisons on NAFNet \cite{chen2022simple}, AIR finetuned i) with latent diffusion weight (AIR-v0) and ii) with weight pre-trained on seven tasks (AIR-v1) on unseen de-snowing and JPEG task.
}
    \label{fig:foundation}
\end{figure}

\begin{table}[t]
    \centering
\vspace{0cm}
\setlength{\tabcolsep}{3pt}
\captionof{table}{ Ablation study on BIQA and SCM on deraindrop and dehazing tasks.}
\label{tab:ablation}
\centering
\footnotesize
\def\arraystretch{1.3}
\scalebox{0.8}{
\begin{tabular}{cccccccc}\toprule
  \multicolumn{2}{c}{} & \multicolumn{3}{c}{\textbf{Deraindrop}} & \multicolumn{3}{c}{\textbf{Dehaze}} \\
  \scriptsize BIQA & \scriptsize SCM
   &  \scriptsize SSIM $\uparrow$ & \scriptsize PSNR $\uparrow$ & \scriptsize LPIPS$\downarrow$ & \scriptsize SSIM$\uparrow$ & \scriptsize PSNR$\uparrow$ & \scriptsize LPIPS$\downarrow$ \\
   \cmidrule(l){1-2} \cmidrule(l){3-5} \cmidrule(l){6-8} 
   \ding{55} & \ding{55} & 0.738 & 24.84 & 0.104 & 0.763 &23.49 &0.091 \\
  \ding{55} & \checkmark & 0.903 & {28.84} & {0.068} & 0.925 &  27.53 &  0.038 \\
   \checkmark & \ding{55}  &0.896 & 25.58 &  0.092 & 0.804 & 24.47 &  0.074\\
   \checkmark & \checkmark &\textbf{0.924} &\textbf{30.10} &\textbf{0.063}  &\textbf{0.973} &\textbf{29.34} &\textbf{0.020}\\
\bottomrule
\end{tabular}}
\label{fig:foundation}
\end{table}
\subsection{Ablation Studies} 
\label{subsec:ablation}
The first ablation study is to validate the importance of the Semantic-Agnostic 
BIQA and Structural Correction Module (SCM), as shown in Tab.~\ref{tab:ablation}.
The presence of proper guidance from the SA-BIQA is crucial for enabling accurate image restoration or enhancement in complex tasks. As shown in Tab.~\ref{tab:ablation}, incorporating the  SA-BIQA into the AutoDIR pipeline consistently improves performance in multitasking scenarios. 
Similarly, the Structural-Correction latent diffusion can maintain fine image structures compared with original latent-diffusion-based image restoration. Tab.~\ref{tab:ablation} and Fig.~\ref{fig:SCM} also showed that incorporating the SCM consistently achieves performance gains.

Our second ablation study is to investigate the possibility of using the all-in-one image restoration backbone (AIR) as a foundation model for general image restoration, which is based on the hypothesis that joint learning of multiple image restoration tasks may result in basis operators that are useful for other image restoration tasks. 
%
%
Inspired by prior work on vision foundation models~\cite{wang2022ofa, zhu2022uni,singh2022flava}, we conduct experiments on the unseen de-snowing and JPEG tasks to evaluate this hypothesis.
For our de-snowing experiments, we utilize 90 snowy-clean image pairs of the Comprehensive Snow Dataset (CSD)~\cite{chen2021all} to fine-tune the AIR which is initialized with the pre-trained AIR weights obtained from training on the seven tasks. For the JPEG compression task, we use 40 compressed-clean image pairs of LIVE dataset~\cite{sheikh2006statistical}.
Fig.~\ref{fig:foundation} demonstrates the fine-tuned AIR with the initialization of pre-trained weight adapts to the de-snowing task more quickly compared to the module without the pre-trained weights. 
This shows the new image restoration task can benefit from the knowledge acquired from other image restoration tasks, indicating the implicit correlations between them.
Moreover, it also outperforms the NAFNet specifically trained for the de-snowing task, which further shows the feasibility of constructing a foundational image restoration model by exploring the underlying correlations.

%
%
%
%

\section{Conclusions and Discussions}
\label{sec:conclusion}

In this paper, we present AutoDIR, an all-in-one image restoration system with latent diffusion, which is capable of automatically restoring images with multiple unknown degradations and also supports open-vocabulary image editing user instructions.
By jointly training from multiple individual image restoration datasets, AutoDIR can be viewed as a foundation model for image restoration, which can automatically detect and address multiple image degradations (with SA-BIQA) and enables effective few-shot learning for novel tasks. 
%

 %
 %
To our knowledge, AutoDIR is the most general comprehensive framework for automatic image restoration in the context of complex real-scenario images with multiple unknown degradations. 
We hope that this framework can provide a solid foundation for further works on complex real-scenario images and has the potential to be extended to various practical applications, such as local image enhancement and multi-task video restoration.

There are several limitations of AutoDIR that we plan to address in the future. First, the computational cost is similar to Latent Diffusion~\cite{rombach2022high} and other diffusion-based image restoration methods~\cite{wang2023stableSR}, which is often significantly higher than non-generative networks such as NAFNet~\cite{chen2022simple}. Some recent work on accelerating diffusion networks such as one-step diffusion~\cite{liu2023insta} provides possible solutions. Second, AutoDIR currently focuses more on global image restoration, such as deblurring, denoising, and low-light enhancement,  rather than local region-based image editing, such as flare removal and inpainting. Extending AutoDIR for local image editing by incorporating semantic segmentation may be an interesting direction to explore.

\clearpage
\newpage

\clearpage
\setcounter{page}{1}

\appendix
\begin{figure}[hpt!]
    \centering
\includegraphics[width=\textwidth]{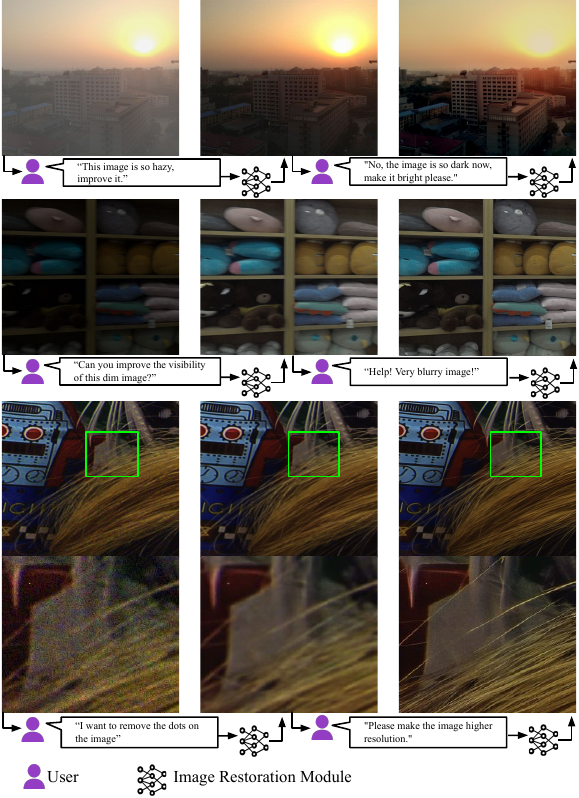}
    \caption{User specified results. Users can edit the image according to their preference via \textbf{open-vocabulary} text instructions. The first column: original image $I_{0}$. The second column: the edited image $I_{1}$ after the user's first instruction given image $I_{0}$. The third column: the second edited image $I_{2}$ after the user's second instruction given the image $I_{1}$.}
    \label{fig:user}
\end{figure}
\begin{figure*}[hpt!]
    \centering
\includegraphics[width=\textwidth]{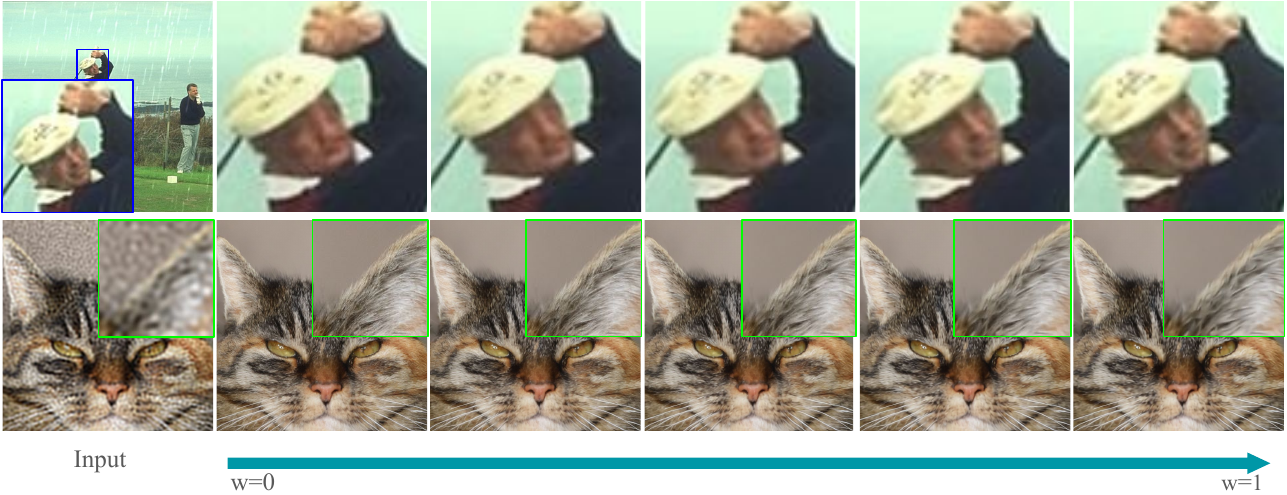}
    \caption{Trade-off of the adjustable coefficient $w$ for structural-correction model. The first row demonstrates that large $w$ can recover the structural details of the original image. Conversely, the second row shows that a smaller $w$ can maintain the generation capability of the generative latent diffusion model.}
    \label{fig:change_w}
\end{figure*}

\begin{figure*}[hpt!]
    \centering
\includegraphics[width=\textwidth]{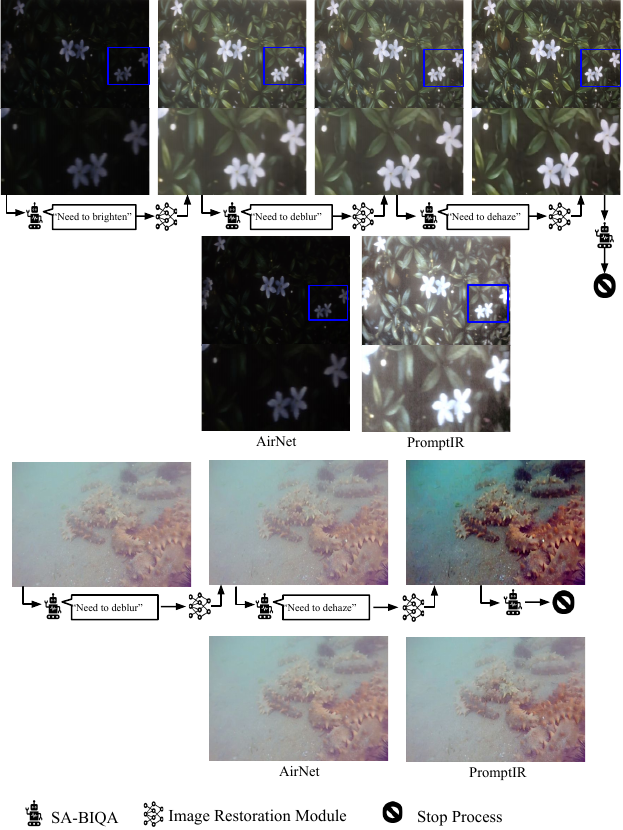}
    \caption{Handling images of multiple \textbf{unknown} artifacts in the \textbf{unseen real-world} datasets.}
    \label{fig:multi-step1}
\end{figure*}

\begin{figure*}[hpt!]
    \centering
\includegraphics[width=1.05\textwidth]{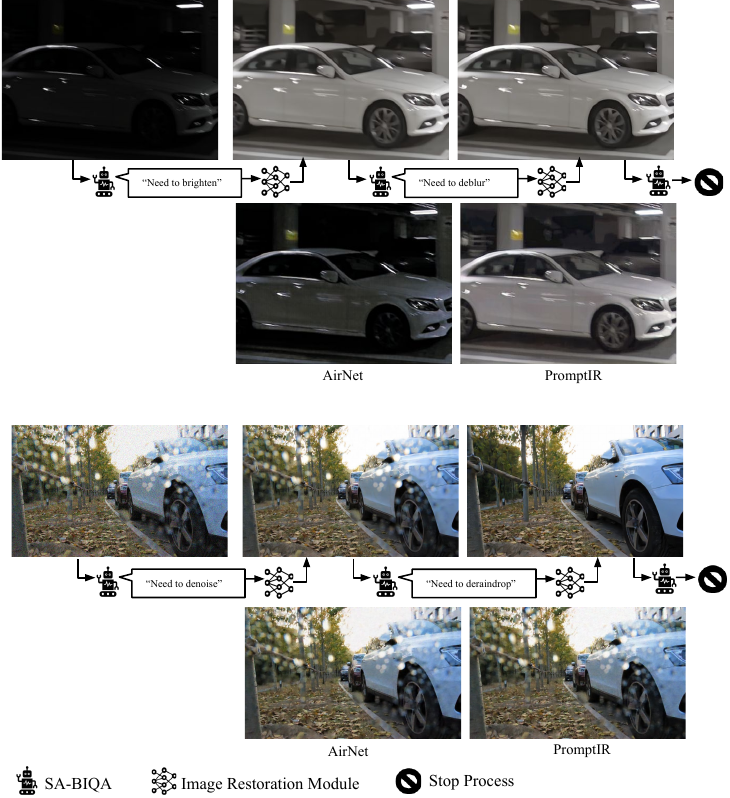}
    \caption{Handling images of multiple \textbf{unknown} artifacts in the \textbf{unseen real-world} datasets.}
    \label{fig:multi-step2}
\end{figure*}

\begin{figure*}[hpt!]
    \centering
\includegraphics[width=0.9\textwidth]{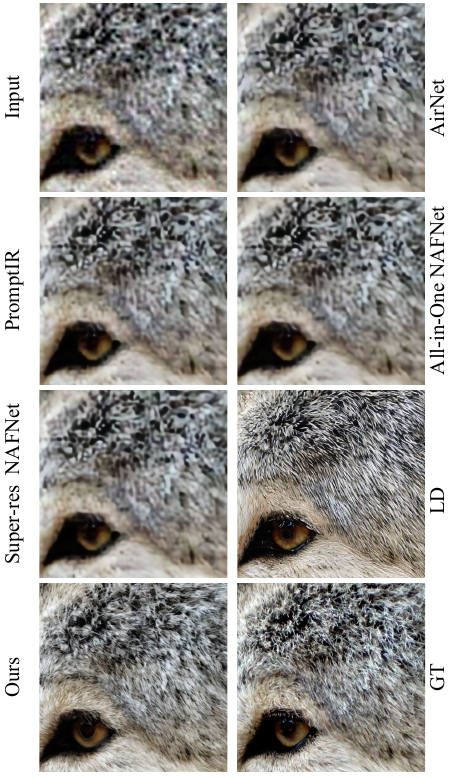}
    \caption{Qualitative comparisons on Super-Resolution.}
    \label{fig:sr_sp_1}
\end{figure*}

\begin{figure*}[hpt!]
    \centering
\includegraphics[width=0.9\textwidth]{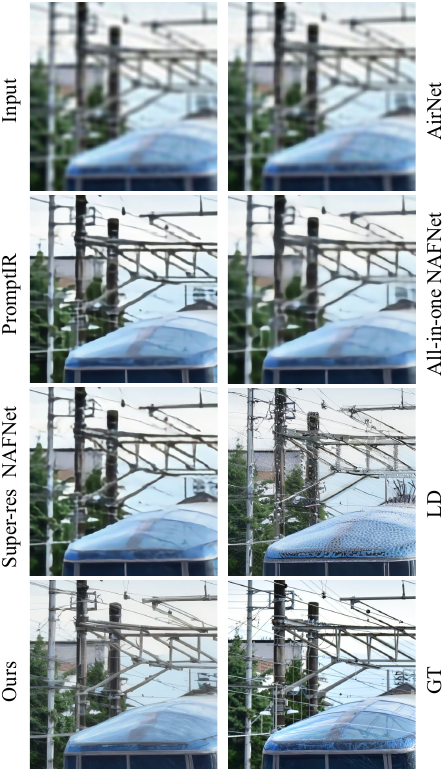}
    \caption{Qualitative comparisons on Super-Resolution.}
    \label{fig:sr_sp_2}
\end{figure*}

\begin{figure*}[hpt!]
    \centering
\includegraphics[width=0.9\textwidth]{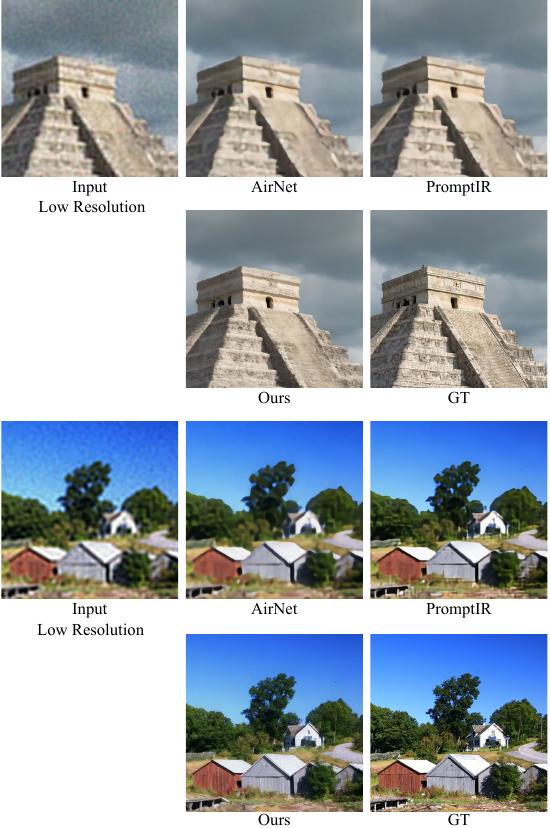}
    \caption{Qualitative comparisons on Super-Resolution with state-of-the-art all-in-one methods.}
    \label{fig:sp_more1}
\end{figure*}

\begin{figure*}[hpt!]
    \centering
\includegraphics[width=0.9\textwidth]{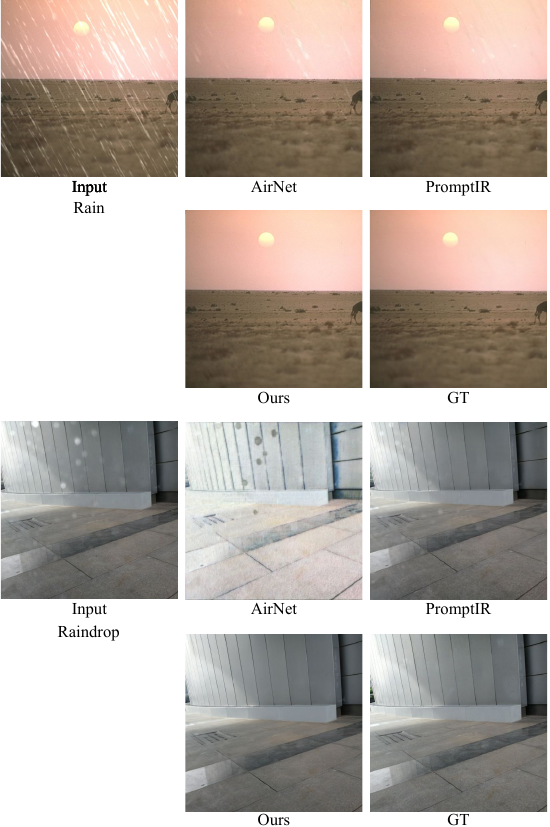}
    \caption{Qualitative comparisons on derain and deraindrop with state-of-the-art all-in-one methods.}
    \label{fig:sp_more2}
\end{figure*}

\begin{figure*}[hpt!]
    \centering
\includegraphics[width=0.9\textwidth]{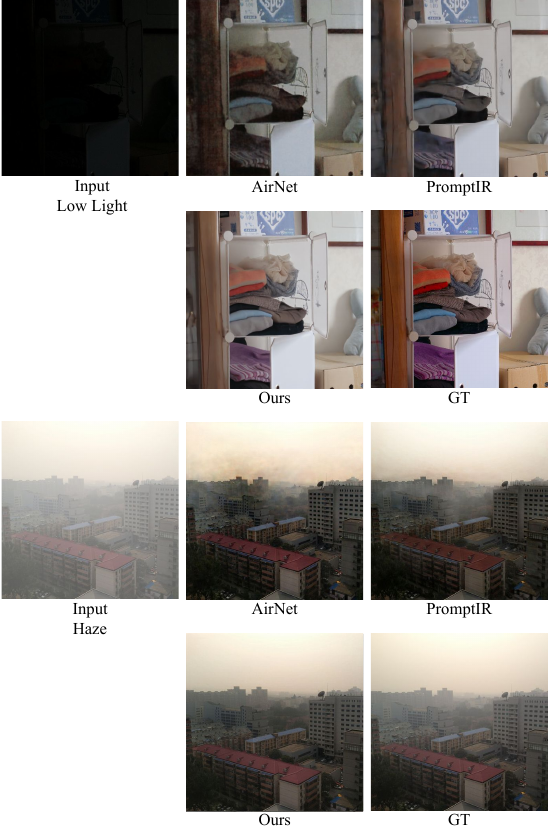}
    \caption{Qualitative comparisons on low light enhancement and dehazing with state-of-the-art all-in-one methods.}
    \label{fig:sp_more3}
\end{figure*}

\begin{figure*}[hpt!]
    \centering
\includegraphics[width=0.9\textwidth]{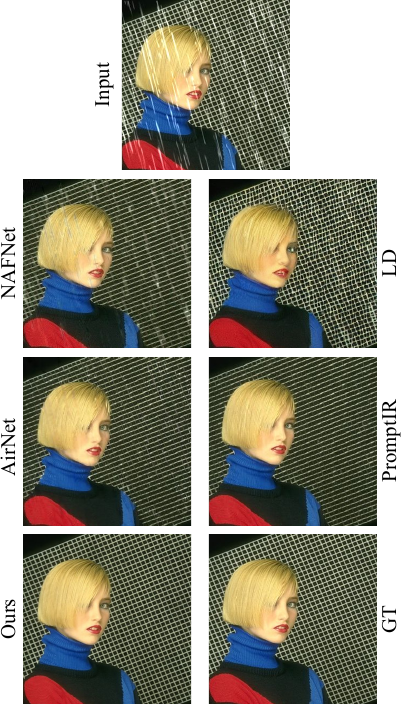}
    \caption{Zoomed-in deraining results in the main text.}
    \label{fig:sp_big1}
\end{figure*}

\begin{figure*}[hpt!]
    \centering
\includegraphics[width=0.9\textwidth]{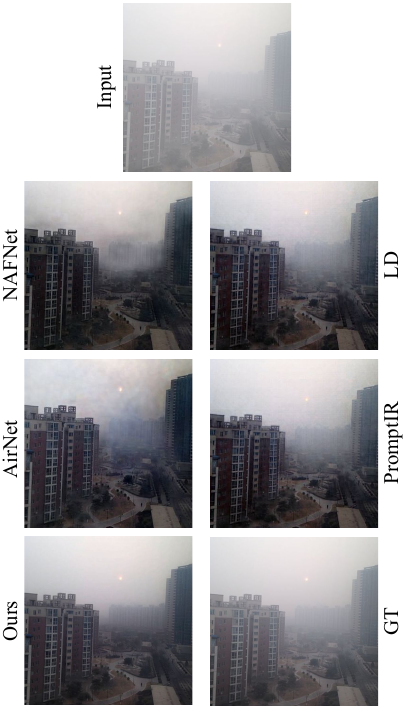}
    \caption{Zoomed-in dehazing results in the main text.}
    \label{fig:sp_big2}
\end{figure*}

\begin{figure*}[hpt!]
    \centering
\includegraphics[width=0.9\textwidth]{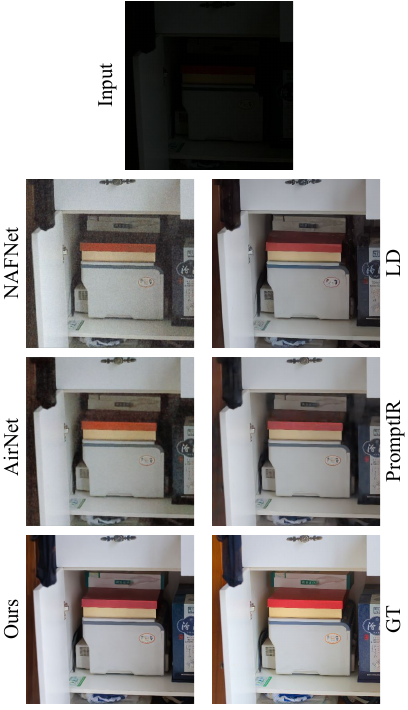}
    \caption{Zoomed-in low light enhancement results in the main text.}
    \label{fig:sp_big3}
\end{figure*}

\begin{figure*}[hpt!]
    \centering
\includegraphics[width=0.9\textwidth]{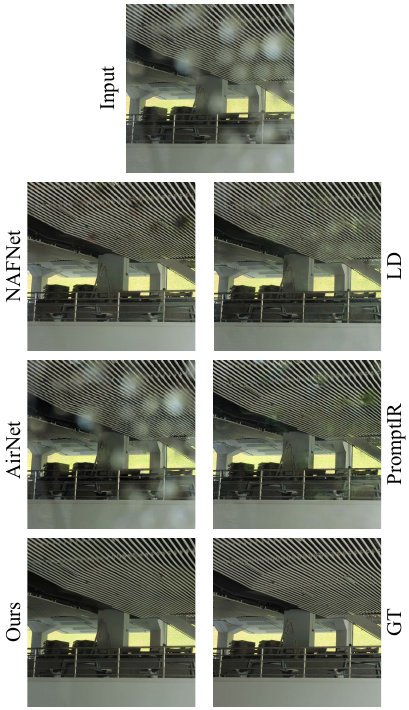}
    \caption{Zoomed-in deraindrop results in the main text.}
    \label{fig:sp_big4}
\end{figure*}

\begin{figure*}[hpt!]
    \centering
\includegraphics[width=0.9\textwidth]{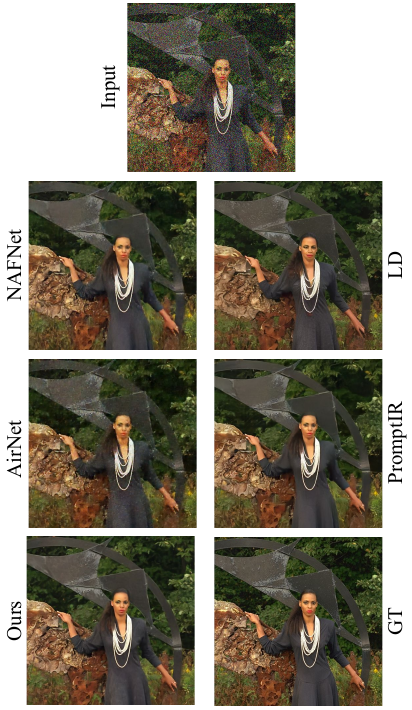}
    \caption{Zoomed-in denoise results in the main text.}
    \label{fig:sp_big5}
\end{figure*}

\begin{figure*}[hpt!]
    \centering
\includegraphics[width=0.9\textwidth]{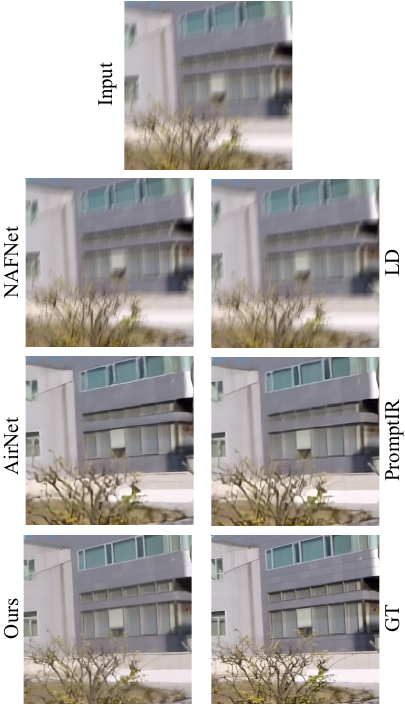}
    \caption{Zoomed-in deblur results in the main text.}
    \label{fig:sp_big6}
\end{figure*}

\noindent This supplementary is organized as follows:
\begin{itemize}
    \item In Sec. \ref{sec:user_specify}, we first introduce our \textbf{open-vocabulary} user interaction application, highlighting its significance in enhancing user engagement. 
    \item Next, in Sec. \ref{sec:trade-off}, we discuss the trade-off of selecting adjustable coefficient $w$ introduced in Sec.~3 of the main body.
    \item To address the restoration of real-world images affected by multiple unknown degradations, we present further results in Sec. \ref{sec:multiple_unknown}.
    \item To further validate the effectiveness of our proposed approach, we present additional qualitative results in Sec. \ref{sec:seven_tasks}, including seven image restoration tasks: denoising, super-resolution, deblurring, deraining, dehazing, low light enhancement, and deraindrop.
    \item In Sec. \ref{subsec:ablation_studies}, we conduct additional ablation studies focusing on the semantic-agnostic constraint and the structural-correction module, providing deeper insights into their contributions.
    \item In Sec. \ref{sec:implementation_details}, we provide full implementation details of our methods and corresponding experiments to  for reproducing our results. 
    \item In Sec. \ref{sec:user_study}, we conduct a user study specifically targeting the real-world multiple degradation image restoration task. This study aims to provide additional evidence regarding the perceptual quality and effectiveness of AutoDIR in comparison to alternative methods.
    \item In Sec. \ref{sec: real_SR_denoise}, we demonstrate the effectiveness of AutoDIR on real-world unseen Super-resolution and Denoise datasets. 
    \item In Sec. \ref{sec: real_SA-BIQA}, we evaluate SA-BIQA on real-world unseen datasets to illustrate the robustness of SA-BIQA. 

\end{itemize}

\section{Open-Vocabulary User Interaction }
\label{sec:user_specify}
As shown in Fig.~\ref{fig:user}, AutoDIR provides a customizable approach to tailor the result outputs based on user preferences. Users can effectively modify the input image by providing corresponding \textbf{open-vocabulary} text prompts. 
The support of user interaction highlights the flexibility and adaptability of our proposed approach, allowing for a highly customizable image enhancement experience. 

\section{Trade-off of adjustable coefficient $w$}
\label{sec:trade-off}
The value of $w$ for the weight of the structural correction module introduced in Sec.~3 of the main body determines the extent to which contextual information is utilized to recover the final result. Fig.~\ref{fig:change_w} demonstrates that a larger value of $w$ helps to recover the complex structures e.g. human face, of the original image.  On the opposite, for tasks like super-resolution, a smaller value of $w$ is required to maintain the generation capability of the latent diffusion model. 

\section{Extensive Experimental Results}
This section provides additional visualization and experimental details on datasets and training settings in the main text's Sec.~4. 

\subsection{Results on Images with Multiple Unknown Degradations in Unseen Real-world Datasets}
\label{sec:multiple_unknown}
We present additional visualization results in Fig. \ref{fig:multi-step1} and Fig. \ref{fig:multi-step2} on images with multiple unknown degradations on \textbf{unseen real-world} UCD~\cite{zhou2020image}, EVUP~\cite{islam2020fast}, LOL-Blur~\cite{zhou2022lednet} and RainDS~\cite{quan2021removing} datasets, to further demonstrate the performance of our method in handling such complex scenarios.

\subsection{Results on Seven Joint-learned Tasks}
\label{sec:seven_tasks}
We provide more visualization results on the seven image restoration tasks in Fig.~\ref{fig:sr_sp_1},~\ref{fig:sr_sp_2}, ~\ref{fig:sp_more1},~\ref{fig:sp_more2},~\ref{fig:sp_more3} and we also provide the zoom-in visualization results in the main text in Fig.~\ref{fig:sp_big1},~\ref{fig:sp_big2},~\ref{fig:sp_big3},~\ref{fig:sp_big4}, ~\ref{fig:sp_big5},~\ref{fig:sp_big6}. 
\begin{figure}[h!]
    \centering
\includegraphics[width=0.6\columnwidth]{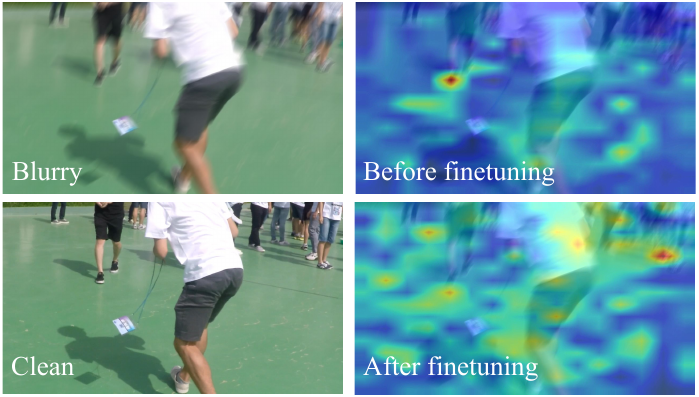}
    \caption{\textbf{Self-attention maps of SA-BIQA image encoder.} Attention maps focus on (have the highest attention values) the foreground object in BIQA image encoder without fine-tuning, while attention maps focus on (have the highest attention values) both foreground and background objects in SA-BIQA image encoder fine-tuned with SA constraint.
}
    \label{fig:clip_att}

\end{figure}

\subsection{Ablation Studies}
\label{subsec:ablation_studies}


\paragraph{\textbf{Fine-tuning with Semantic-agnostic Constraint for BIQA}} 

In order to further analyze the effectiveness of our semantic-agnostic constraint, we visualize the attention map of the image encoder in Fig. \ref{fig:clip_att}.
As depicted in the visualization, before fine-tuning, the attention map primarily focuses on the pronominal object, which is consistent with the behavior of the original CLIP model \cite{radford2021learning} that was pre-trained on image classification tasks.
%
%
After fine-tuning with semantic-agnostic constraint, we observe that the attention maps expand to highlight background areas that may contain potential artifacts which shows that the BIQA model has successfully learned to prioritize and focus on artifacts, leading to more accurate BIQA results.

\paragraph{\textbf{Importance of Structural-Correction Module (SCM) for latent diffusion}}
The generative Latent Diffusion model \cite{rombach2022high} has demonstrated a strong ability to generate unseen features. However, it falls short when preserving the original structural information of the input image, which is crucial for image enhancement tasks.

\begin{figure*}[hpt!]
    \centering
\includegraphics[width=\textwidth]{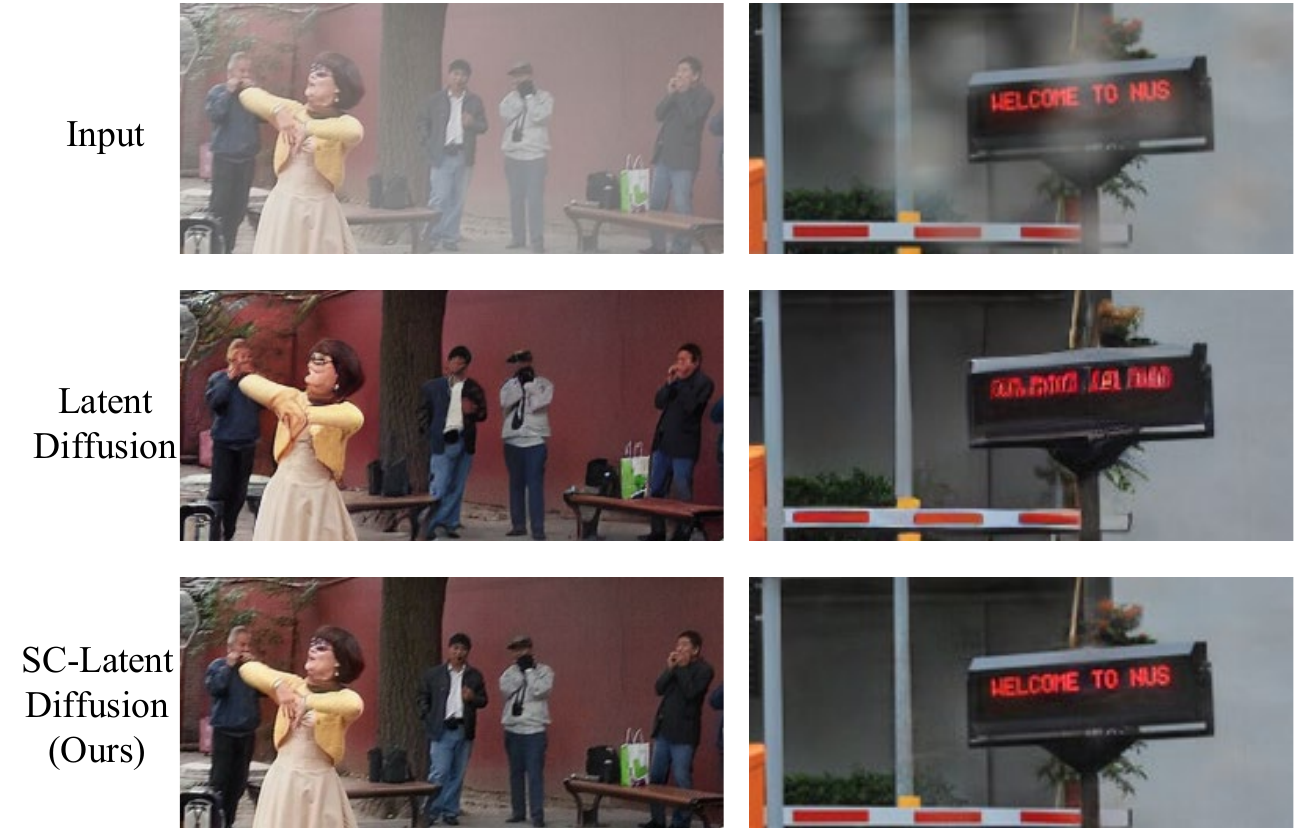}
    \caption{Qualitative comparisons for latent diffusion model and LC-latent diffusion (ours) on dehazing and deraindrop tasks. 
}
    \label{fig:scm}
\end{figure*}
\begin{figure*}[hpt!]
    \centering
\includegraphics[width=\textwidth]{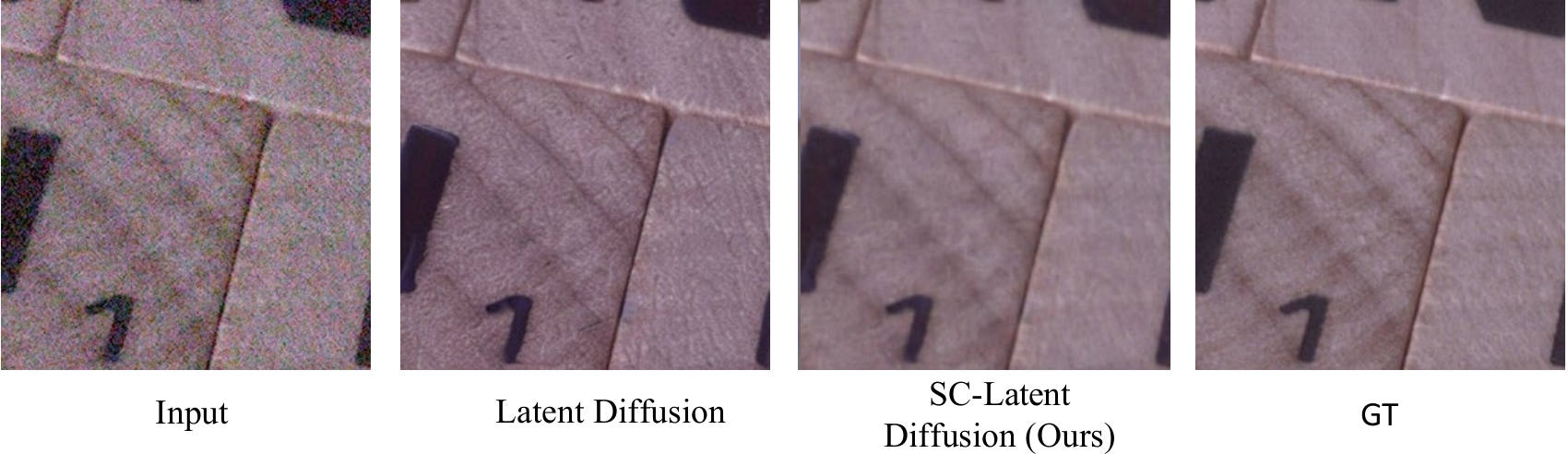}
    \caption{Qualitative comparisons for latent diffusion model and LC-latent diffusion (ours) on denoising tasks. 
}
    \label{fig:scm_denoising}
\end{figure*}

As illustrated in Fig.~\ref{fig:scm}, Structural-Correction Latent-Diffusion has high-quality results with fine details intact. On the other hand, latent diffusion exhibits significant distortion in faces and text. Moreover, Fig.~\ref{fig:scm_denoising} demonstrates that the structural-correction module also shows the ability to correct hallucinated undesire textures of the results of the latent diffusion.

\paragraph{\textbf{BIQA performance comparison of ViT and CLIP-variants.}}
As shown in Tab.~\ref{tab:precision}, we report the F-scores of seven degradation tasks, our SA-CLIP outperforms ViT Classifier, pre-trained CLIP, and naively finetuned CLIP in all the seven tasks.

\begin{table}[ht!]
		\setlength{\abovecaptionskip}{0pt} \setlength{\belowcaptionskip}{0pt}
\caption{\small F-Score of image degradation detection on seven degradation tasks and clean image.}
\label{tab:precision}
\centering
\footnotesize
\def\arraystretch{1.4}
\scalebox{0.9}{
\begin{tabular}{cccccccc}\toprule

  Degradation &  Haze & Blur & Rain &LOL &Raindrop &Noise &Low-Res\\\midrule
   ViT-classifier & 0.6666 & 0.7714 & 0.9603 &0.7692 &1.0000 & 0.6934 & 0.8450
\\
   Original CLIP & 0.6738 & 0.7206 & 0.8889 &0.8823 &0.7568 &0.8049 &0.7229
\\
   Fine-tuned CLIP &  0.6955 & 0.7744 &0.9135 &0.9091 & 0.9464 & 0.7463 &0.9085
\\
   \textbf{ SA-CLIP (ours) } & \textbf{1.0000} & \textbf{0.9444} & \textbf{0.9950} & \textbf{1.0000} &\textbf{1.0000} &\textbf{1.0000} & \textbf{0.9985}
\\
  \bottomrule
\end{tabular}}
\end{table}

\section{Implementation Details}
\label{sec:implementation_details}
Our experiments are conducted using PyTorch on a computational setup comprising eight NVIDIA V100 GPUs. The training process for the AutoDIR framework involves three distinct steps.
Firstly, we initialize the training by freezing the text encoder and fine-tuning the image encoder within the Blind Image Quality Assessment (BIQA). We utilize the Adam optimizer with a batch size of 1024 and train for 20 epochs. The initial learning rate is set to 3 $\times 10^{-6}$ and follows a cosine annealing rule.
Next, we proceed to fine-tune the All-in-One Image Restoration (AIR) backbone using the Adam optimizer. During this stage, we employ a learning rate of $1e^{-4}$ and a batch size of 256. The fine-tuning process is performed for 15 epochs.
Finally, we freeze the previously trained pipeline components and focus on training the structural-correction module (SCM). For this stage, we employ the Adam optimizer and a cosine annealing rule. The initial learning rate is set to $1e^{-3}$, and the batch size is 256. We train the SCM for 8,000 iterations.
In our experiments, the structural-correction module for Structural-Correction-Latent Diffusion is based on NAFNet architecture \cite{chen2022simple}.
During inference, the coefficient $w$ of the structural-correction module is set to be 1 as default for denoising, deraining, dehazing, deraindrop, low light enhancement, and debluring tasks and 0.1 for the super-resolution task to maintain the generation capability of the generative latent diffusion model.

\paragraph{\textbf{Datasets}:}
The seven image restoration tasks are denoising, deblurring, super-resolution, low-light enhancement, dehazing, deraining, and deraindrop. 
For denoising, we use SIDD~\cite{SIDD_2018_CVPR} and a synthetic Gaussian and Poisson noise dataset with DIV2K~\cite{agustsson2017ntire} and Flickr2K~\cite{lim2017enhanced}. 
For super-resolution, we follow previous practice and train AutoDIR with DIV2K~\cite{agustsson2017ntire} and Flickr2K~\cite{lim2017enhanced} training sets, following~\cite{wang2021real} for degraded image generation. 
In addition, we use GoPro~\cite{nah2017deep}, LOL~\cite{wei2018deep}, RESIDE~\cite{li2018benchmarking}, Rain200L~\cite{yang2017deep}, and Raindrop~\cite{qian2018attentive} for deblurring, low-light enhancement, dehazing, deraining, and deraindrop, respectively.
During inference, we evaluate multiple test sets. These include SIDD~\cite{SIDD_2018_CVPR}, Kodak24~\cite{kodak}, DIV2K~\cite{agustsson2017ntire}, GoPro~\cite{nah2017deep}, LOL~\cite{wei2018deep}, SOTS-Outdoor~\cite{li2018benchmarking}, Rain100~\cite{yang2017deep}, and Raindrop~\cite{qian2018attentive}, each corresponding to their respective tasks.
For experiments with unknown degradations, we use the Under-Display Camera (TOLED) dataset~\cite{zhou2021image} and the Enhancing Underwater Visual Perception (EUVP) dataset~\cite{islam2020fast}.

\section{User Study}
\label{sec:user_study}
\begin{figure*}[h]
    \centering
\includegraphics[width=\textwidth]{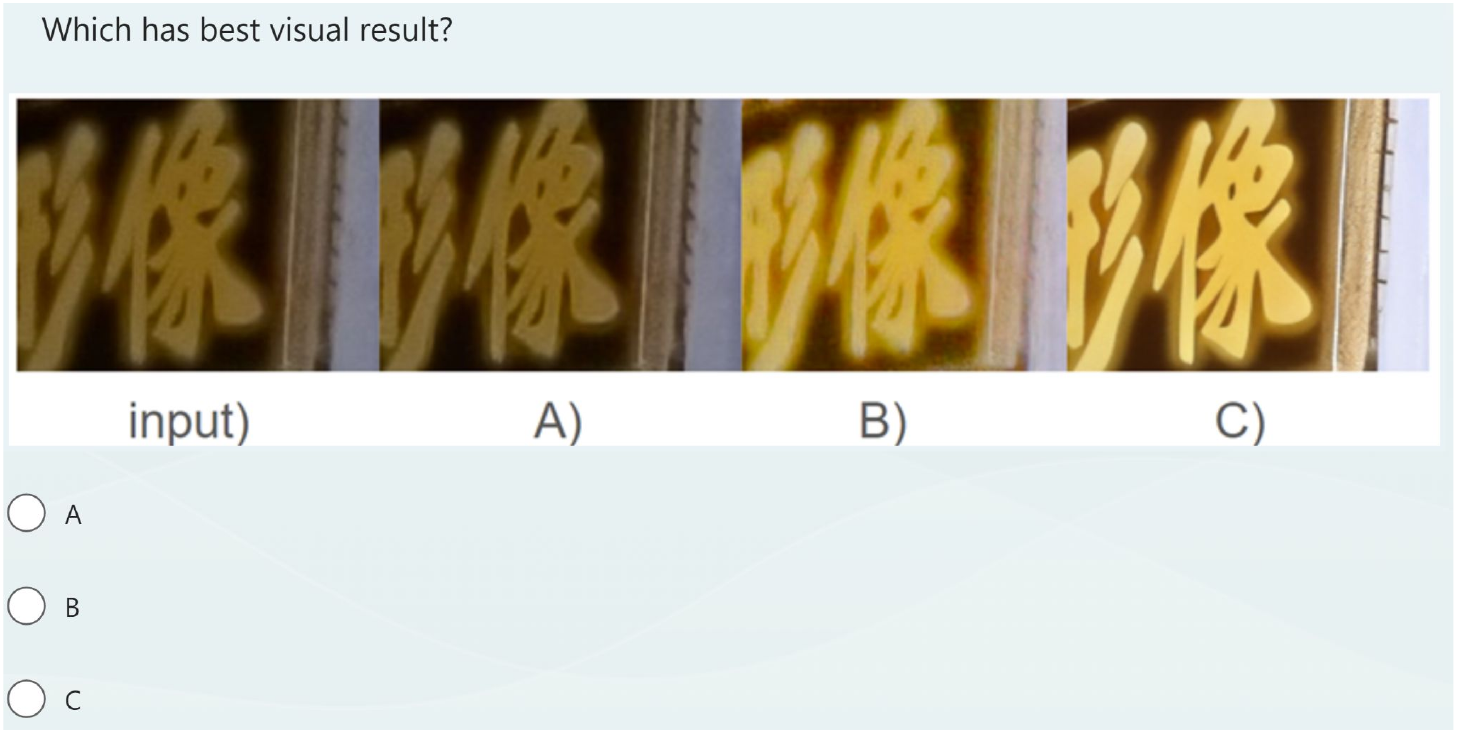}
    \caption{Screenshot of the user interface in the user study.
}
    \label{fig:user_study}
\end{figure*}

\begin{figure}[htp!]
    \centering
\includegraphics[width=0.6\textwidth]{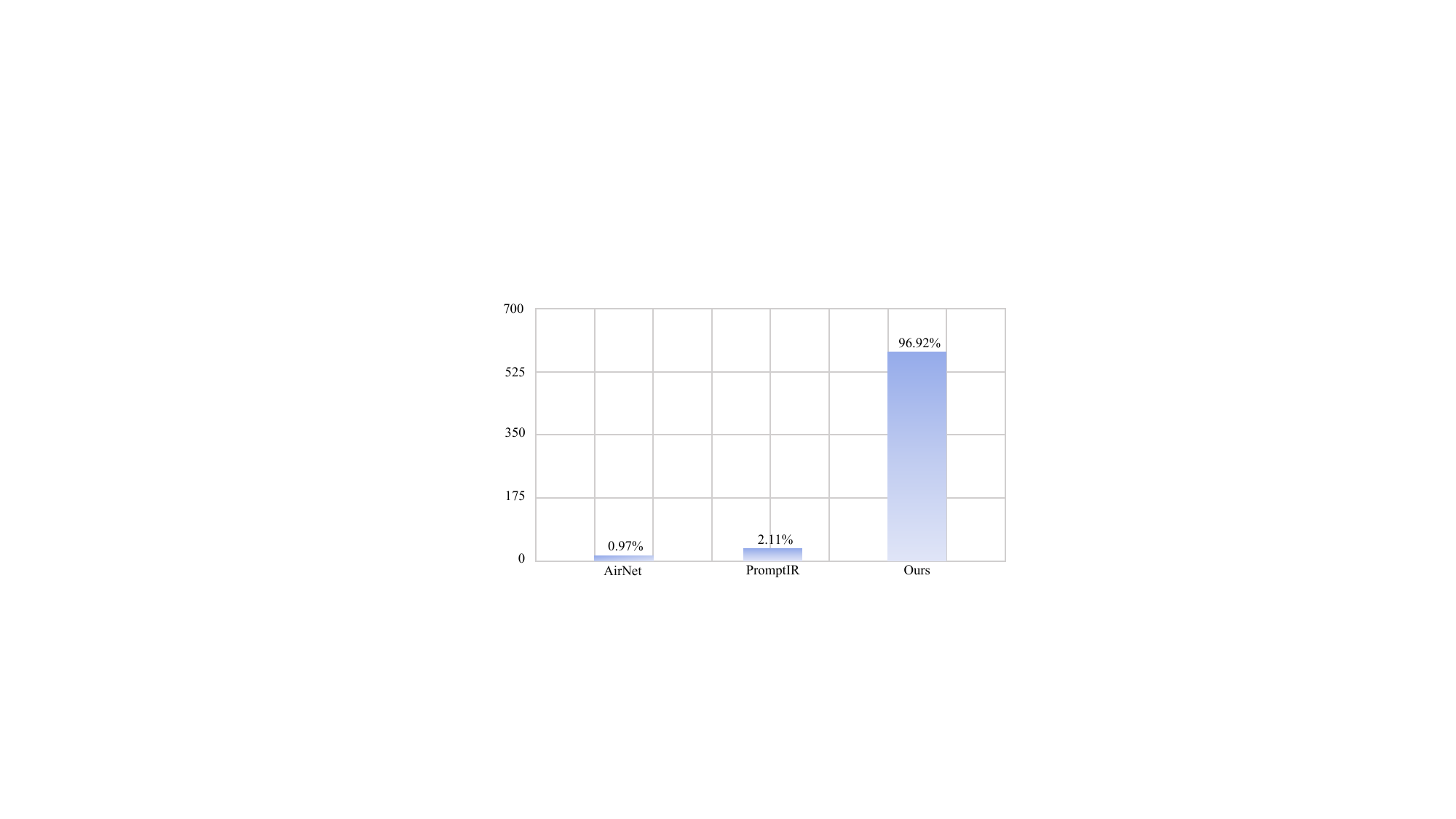}
    \caption{Results of the user study on 28 real-world images with unknown artifacts on unseen datasets or real-captured images. We collected 22 forms and 28 $\times$ 22 $=$ 616 responses in total. Among them, AutoDIR receives more than $96\%$ of the votes as the best results.
}
    \label{fig:user_study_statistic}
\end{figure}

To further examine the effectiveness of AutoDIR, we conduct a user study on the images with unknown degradations in unseen real-world datasets or real-captured images. We compare AutoDIR with state-of-the-art all-in-one AirNet \cite{AirNet} and PromptIR \cite{potlapalli2023promptir}. As shown in Fig.~\ref{fig:user_study}, given the input and the restored results, the question is to ask which image has the best visual, and the choices are in random order. We collect 22 forms and there are $22 \times 28 = 616$ responses in total. Fig.~\ref{fig:user_study_statistic} illustrates that AutoDIR gathers more than $96\%$ of the votes for producing the best denoising results.


\section{Comparison on unseen real-world Super-res and Denoise dataset}
\label{sec: real_SR_denoise}
\begin{figure}[h!]
     \vspace{-2pt}
    \centering
    \vspace{-10pt}\includegraphics[width=1.0\linewidth]{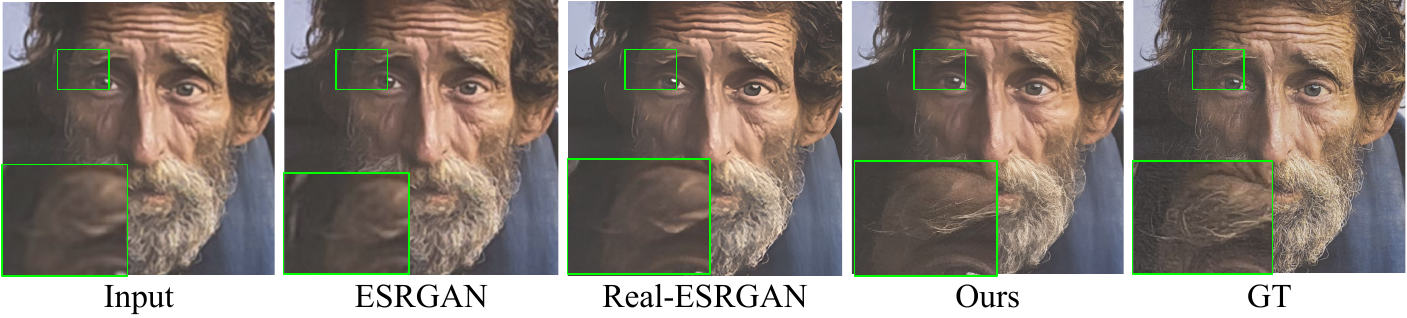}
     \caption{Comparision on RealSR dataset with task-specific methods ESRGAN~\cite{wang2018esrgan} and Real-ESRGAN~\cite{wang2021real}.
}
    \label{fig:realsr}
\end{figure}

 We conduct experiments on the unseen real-world Super-Res dataset (RealSR test dataset \cite{Ji_2020_CVPR_Workshops}) without specific fine-tuning. As illustrated in Fig.~\ref{fig:realsr}, AutoDIR successfully reconstructs details such as eyebrows and beards, which other methods struggle to achieve. Additionally, we have included the quantitative results in Tab.~\ref{tab:RealSR}.

\begin{figure}[h]
    \centering
    \vspace{-10pt}
\includegraphics[width=1.0\linewidth]{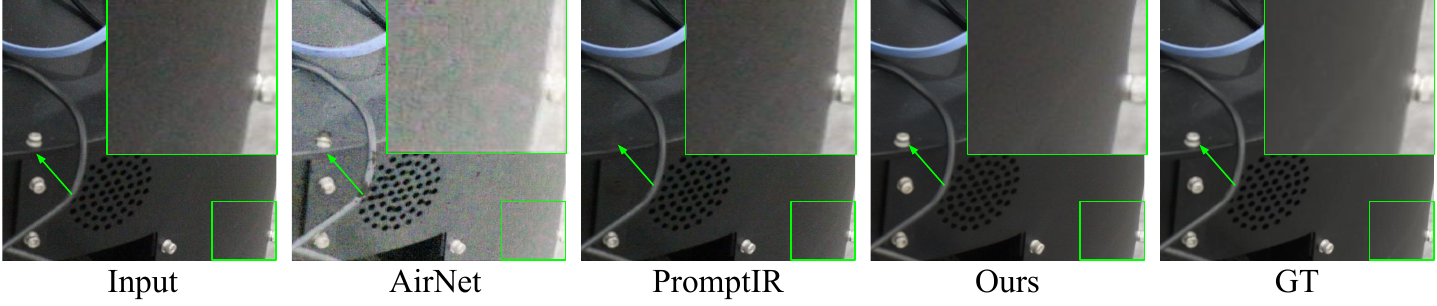}
\caption{Comparision on PolyU-Denoise dataset with all-in-one methods.}
    \label{fig:realnoise}
\end{figure}

As shown in Fig~\ref{fig:realnoise} and Tab.~\ref{tab:RealNoise}, we have also presented quantitative and qualitative results on the unseen real-world denoise dataset (PolyU-Denoise~\cite{xu2018real}), demonstrating the benefits of the AutoDIR approach.

\begin{minipage}{0.56\textwidth}
\captionof{table}{\small Quantitative comparison on RealSR}
\def\arraystretch{1.0}
\scalebox{0.8}{
\begin{tabular}{ccccc}\toprule
  \multirow{2}{*}{\textbf{Method}}   & \multicolumn{4}{c}{\textbf{Real-World Super Resolution}}\\
  &\scriptsize MUSIQ $\uparrow$ & \scriptsize CLIP-IQA $\uparrow$ & \scriptsize NIQE $\downarrow$ & \scriptsize NIMA $\uparrow$  \\
    \midrule
   NAFNet-SR   & 40.53 & 0.251& 7.222 & 4.247
\\
   AirNet   &21.73 & 0.239 & 11.839 & 3.773  \\
   
   PromptIR & 24.85 & 0.248 &8.526 & 4.082  \\
   
   ESRGAN & 30.10 & 0.231&7.819 & 3.942  \\

Real-ESRGAN+ &\underline{60.11} & \underline{0.462} & \textbf{5.130} & \underline{4.660}\\
  \textbf{Ours} &\textbf{60.14} & \textbf{0.493} &\underline{5.204} & \textbf{4.717}\\
  \bottomrule
\end{tabular}}
\label{tab:RealSR}
\end{minipage}
\hfill
\begin{minipage}{0.4\textwidth}
\captionof{table}{\small Quantitative comparison on PolyU-Denoise}
\setlength{\tabcolsep}{2pt}
\scalebox{0.8}{
\begin{tabular}{cccc} \toprule
  \multirow{2}{*}{\textbf{Method}}   & \multicolumn{3}{c}{\textbf{Real-world Denoise}}\\
  &\scriptsize SSIM$\uparrow$ & \scriptsize PSNR$\uparrow$ & \scriptsize LPIPS$\downarrow$   \\
    \midrule
   NAFNet   &0.924 & 34.16&\underline{0.130}
\\
   AirNet   &0.783& 26.26 & 0.293  \\
   
   PromptIR &\underline{0.933} & \underline{34.43} &0.176  \\
   
   LD &  0.910 & 32.85&0.196  \\

  \textbf{Ours} &\textbf{0.942} & \textbf{36.18} &\textbf{0.129}\\
  \bottomrule
\end{tabular} }
\label{tab:RealNoise}

\end{minipage}

\section{Evaluation of SA-BIQA on unseen real-world datasets.}
\label{sec: real_SA-BIQA}

We have conducted experiments on unseen real-world datasets, including Super-Res (RealSR~\cite{Ji_2020_CVPR_Workshops}), Deblur (RealBlur~\cite{rim_2020_ECCV}), and low-light enhancement (HuaWei~\cite{hai2023r2rnet}). The table below demonstrates that SA-BIQA outperforms other methods by accurately predicting the dominant artifact in all tasks, achieving an accuracy of over $90\%$

\def\arraystretch{1.0}
\begin{table}[ht]
\centering
\vspace{-10pt}
\scalebox{0.9}{
\begin{tabular}{cccc}\toprule
 {\textbf{Method}}  
  & low-res& blur & low-light   \\
    \midrule
    ViT-classifier   & 0.9100 &0.0357 & \underline{0.7666}
\\
   Original CLIP   &0.8500& \underline{0.6704} & 0.2333  \\
   
   Fine-tuned CLIP &\underline{0.9200} & {0.4795} &  0.7000 \\

  \textbf{ SA-CLIP (ours) } &\textbf{1.0000} & \textbf{0.9010} &\textbf{0.9333}\\
  \bottomrule
\end{tabular}}
\end{table}

\clearpage



%
%
{
    \small
    \bibliographystyle{splncs04}
    \bibliography{main}

\begin{thebibliography}{100}
\providecommand{\url}[1]{\texttt{#1}}
\providecommand{\urlprefix}{URL }
\providecommand{\doi}[1]{https://doi.org/#1}

\bibitem{SIDD_2018_CVPR}
Abdelhamed, A., Lin, S., Brown, M.S.: A high-quality denoising dataset for smartphone cameras. In: Proceedings of the IEEE Conference on Computer Vision and Pattern Recognition (CVPR) (2018)

\bibitem{abu2022adir}
Abu-Hussein, S., Tirer, T., Giryes, R.: {ADIR}: Adaptive diffusion for image reconstruction. arXiv preprint arXiv:2212.03221  (2022)

\bibitem{agustsson2017ntire}
Agustsson, E., Timofte, R.: {NTIRE} 2017 challenge on single image super-resolution: Dataset and study. In: Proceedings of the IEEE Conference on Computer Vision and Pattern Recognition (CVPR) Workshops (2017)

\bibitem{priorbau2019semantic}
Bau, D., Strobelt, H., Peebles, W., Wulff, J., Zhou, B., Zhu, J.Y., Torralba, A.: Semantic photo manipulation with a generative image prior. ACM Transactions on Graphics (TOG)  \textbf{38}(4), ~59 (2019)

\bibitem{blau2018perception}
Blau, Y., Michaeli, T.: The perception-distortion tradeoff. In: Proceedings of the IEEE Conference on Computer Vision and Pattern Recognition (CVPR) (2018)

\bibitem{brooks2023instructpix2pix}
Brooks, T., Holynski, A., Efros, A.A.: {InstructPix2Pix}: Learning to follow image editing instructions. In: Proceedings of Advances in Neural Information Processing Systems (NeurIPS) (2023)

\bibitem{priorchan2021glean}
Chan, K.C., Wang, X., Xu, X., Gu, J., Loy, C.C.: {GLEAN}: Generative latent bank for large-factor image super-resolution. In: Proceedings of the IEEE Conference on Computer Vision and Pattern Recognition (CVPR) (2021)

\bibitem{chefer2023attend}
Chefer, H., Alaluf, Y., Vinker, Y., Wolf, L., Cohen-Or, D.: Attend-and-excite: Attention-based semantic guidance for text-to-image diffusion models. ACM Transactions on Graphics (TOG)  \textbf{42}(4),  1--10 (2023)

\bibitem{singletaskchen2022femasr}
Chen, C., Shi, X., Qin, Y., Li, X., Han, X., Yang, T., Guo, S.: Real-world blind super-resolution via feature matching with implicit high-resolution priors. In: IEEE Transactions on Multimedia (TMM) (2022)

\bibitem{chen2022femasr}
Chen, C., Shi, X., Qin, Y., Li, X., Han, X., Yang, T., Guo, S.: Real-world blind super-resolution via feature matching with implicit high-resolution priors (2022)

\bibitem{chen2021pre}
Chen, H., Wang, Y., Guo, T., Xu, C., Deng, Y., Liu, Z., Ma, S., Xu, C., Xu, C., Gao, W.: Pre-trained image processing transformer. In: Proceedings of the IEEE Conference on Computer Vision and Pattern Recognition (CVPR) (2021)

\bibitem{chen2024textdiffuser}
Chen, J., Huang, Y., Lv, T., Cui, L., Chen, Q., Wei, F.: Textdiffuser: Diffusion models as text painters. Proceedings of Advances in Neural Information Processing Systems (NeurIPS)  (2024)

\bibitem{chen2022simple}
Chen, L., Chu, X., Zhang, X., Sun, J.: Simple baselines for image restoration. In: Proceedings of European Conferences on Computer Vision (ECCV) (2022)

\bibitem{chen2021all}
Chen, W.T., Fang, H.Y., Hsieh, C.L., Tsai, C.C., Chen, I., Ding, J.J., Kuo, S.Y., et~al.: All snow removed: Single image desnowing algorithm using hierarchical dual-tree complex wavelet representation and contradict channel loss. In: Proceedings of the IEEE International Conference on Computer Vision (ICCV) (2021)

\bibitem{chen2022learning}
Chen, W.T., Huang, Z.K., Tsai, C.C., Yang, H.H., Ding, J.J., Kuo, S.Y.: Learning multiple adverse weather removal via two-stage knowledge learning and multi-contrastive regularization: Toward a unified model. In: Proceedings of the IEEE Conference on Computer Vision and Pattern Recognition (CVPR) (2022)

\bibitem{cheon2021perceptual}
Cheon, M., Yoon, S.J., Kang, B., Lee, J.: Perceptual image quality assessment with transformers. In: Proceedings of the IEEE Conference on Computer Vision and Pattern Recognition (CVPR) (2021)

\bibitem{dosovitskiy2020image}
Dosovitskiy, A., Beyer, L., Kolesnikov, A., Weissenborn, D., Zhai, X., Unterthiner, T., Dehghani, M., Minderer, M., Heigold, G., Gelly, S., et~al.: An image is worth 16x16 words: Transformers for image recognition at scale. Proceedings of International Conference on Learning Representations (ICLR)  (2020)

\bibitem{fei2023generative}
Fei, B., Lyu, Z., Pan, L., Zhang, J., Yang, W., Luo, T., Zhang, B., Dai, B.: Generative diffusion prior for unified image restoration and enhancement. In: Proceedings of the IEEE Conference on Computer Vision and Pattern Recognition (CVPR) (2023)

\bibitem{kodak}
Franzen, R.: Lossless true color image suite. http://r0k.us/graphics/kodak/  (1999)

\bibitem{goodfellow2014generative}
Goodfellow, I., Pouget-Abadie, J., Mirza, M., Xu, B., Warde-Farley, D., Ozair, S., Courville, A., Bengio, Y.: Generative adversarial nets. Proceedings of Advances in Neural Information Processing Systems (NeurIPS)  (2014)

\bibitem{gu2020image}
Gu, J., Shen, Y., Zhou, B.: Image processing using multi-code {GAN} prior. In: Proceedings of the IEEE Conference on Computer Vision and Pattern Recognition (CVPR) (2020)

\bibitem{hai2023r2rnet}
Hai, J., Xuan, Z., Yang, R., Hao, Y., Zou, F., Lin, F., Han, S.: R2rnet: Low-light image enhancement via real-low to real-normal network. Journal of Visual Communication and Image Representation  (2023)

\bibitem{hertz2022prompt}
Hertz, A., Mokady, R., Tenenbaum, J., Aberman, K., Pritch, Y., Cohen-Or, D.: {Prompt-to-Prompt} image editing with cross attention control. Proceedings of International Conference on Learning Representations (ICLR)  (2023)

\bibitem{ho2020denoising}
Ho, J., Jain, A., Abbeel, P.: Denoising diffusion probabilistic models. Proceedings of Advances in Neural Information Processing Systems (NeurIPS)  (2020)

\bibitem{huang2023collaborative}
Huang, Z., Chan, K.C., Jiang, Y., Liu, Z.: Collaborative diffusion for multi-modal face generation and editing. In: Proceedings of the IEEE Conference on Computer Vision and Pattern Recognition (CVPR) (2023)

\bibitem{priorhussein2020image}
Hussein, S.A., Tirer, T., Giryes, R.: Image-adaptive {GAN} based reconstruction. In: Proceedings of the AAAI Conference on Artificial Intelligence (2020)

\bibitem{islam2020fast}
Islam, M.J., Xia, Y., Sattar, J.: Fast underwater image enhancement for improved visual perception. IEEE Robotics and Automation Letters (RA-L)  (2020)

\bibitem{rim_2020_ECCV}
Jaesung~Rim, Haeyun~Lee, J.W.S.C.: Real-world blur dataset for learning and benchmarking deblurring algorithms. In: Proceedings of European Conferences on Computer Vision (ECCV) (2020)

\bibitem{singletaskji2022xydeblur}
Ji, S.W., Lee, J., Kim, S.W., Hong, J.P., Baek, S.J., Jung, S.W., Ko, S.J.: {XYDeblur}: divide and conquer for single image deblurring. In: Proceedings of the IEEE Conference on Computer Vision and Pattern Recognition (CVPR) (2022)

\bibitem{Ji_2020_CVPR_Workshops}
Ji, X., Cao, Y., Tai, Y., Wang, C., Li, J., Huang, F.: Real-world super-resolution via kernel estimation and noise injection. In: Proceedings of the IEEE Conference on Computer Vision and Pattern Recognition (CVPR) Workshops (2020)

\bibitem{singletaskjiang2020multi}
Jiang, K., Wang, Z., Yi, P., Chen, C., Huang, B., Luo, Y., Ma, J., Jiang, J.: Multi-scale progressive fusion network for single image deraining. In: Proceedings of the IEEE Conference on Computer Vision and Pattern Recognition (CVPR) (2020)

\bibitem{jinjin2020pipal}
Jinjin, G., Haoming, C., Haoyu, C., Xiaoxing, Y., Ren, J.S., Chao, D.: Pipal: a large-scale image quality assessment dataset for perceptual image restoration. In: Proceedings of European Conferences on Computer Vision (ECCV) (2020)

\bibitem{kawar2022denoising}
Kawar, B., Elad, M., Ermon, S., Song, J.: Denoising diffusion restoration models. In: Proceedings of Advances in Neural Information Processing Systems (NeurIPS) (2022)

\bibitem{kawar2023imagic}
Kawar, B., Zada, S., Lang, O., Tov, O., Chang, H., Dekel, T., Mosseri, I., Irani, M.: {Imagic}: Text-based real image editing with diffusion models. In: Proceedings of the IEEE Conference on Computer Vision and Pattern Recognition (CVPR) (2023)

\bibitem{ke2021musiq}
Ke, J., Wang, Q., Wang, Y., Milanfar, P., Yang, F.: {MUSIQ}: Multi-scale image quality transformer. In: Proceedings of the IEEE International Conference on Computer Vision (ICCV)

\bibitem{kingma2013auto}
Kingma, D.P., Welling, M.: Auto-encoding variational bayes (2014)

\bibitem{li2018benchmarking}
Li, B., Ren, W., Fu, D., Tao, D., Feng, D., Zeng, W., Wang, Z.: Benchmarking single-image dehazing and beyond. IEEE Transactions on Image Processing (TIP)  (2018)

\bibitem{AirNet}
Li, B., Liu, X., Hu, P., Wu, Z., Lv, J., Peng, X.: {All-In-One} image restoration for unknown corruption. In: Proceedings of the IEEE Conference on Computer Vision and Pattern Recognition (CVPR) (2022)

\bibitem{li2020all}
Li, R., Tan, R.T., Cheong, L.F.: {All-in-One} bad weather removal using architectural search. In: Proceedings of the IEEE Conference on Computer Vision and Pattern Recognition (CVPR) (2020)

\bibitem{liang2021swinir}
Liang, J., Cao, J., Sun, G., Zhang, K., Van~Gool, L., Timofte, R.: {SwinIR}: Image restoration using swin transformer. In: Proceedings of the IEEE Conference on Computer Vision and Pattern Recognition (CVPR) (2021)

\bibitem{lim2017enhanced}
Lim, B., Son, S., Kim, H., Nah, S., Mu~Lee, K.: Enhanced deep residual networks for single image super-resolution. In: Proceedings of the IEEE Conference on Computer Vision and Pattern Recognition (CVPR) Workshops (2017)

\bibitem{liu2023learning}
Liu, K., Jiang, Y., Choi, I., Gu, J.: Learning image-adaptive codebooks for class-agnostic image restoration. Proceedings of the IEEE International Conference on Computer Vision (ICCV)  (2023)

\bibitem{liu2023insta}
Liu, X., Zhang, X., Ma, J., Peng, J., Liu, Q.: Instaflow: One step is enough for high-quality diffusion-based text-to-image generation. arXiv preprint arXiv:2309.06380  (2023)

\bibitem{priormenon2020pulse}
Menon, S., Damian, A., Hu, S., Ravi, N., Rudin, C.: {PULSE}: Self-supervised photo upsampling via latent space exploration of generative models. In: Proceedings of the IEEE Conference on Computer Vision and Pattern Recognition (CVPR) (2020)

\bibitem{singletaskyao2023csnorm}
Mingde, Y., Jie, H., Xin, J., Ruikang, X., Shenglong, Z., Man, Z., , Zhiwei, X.: Generalized lightness adaptation with channel selective normalization. In: Proceedings of the IEEE International Conference on Computer Vision (ICCV) (2023)

\bibitem{mittal2012making}
Mittal, A., Soundararajan, R., Bovik, A.C.: Making a “completely blind” image quality analyzer. IEEE Signal processing letters (3),  209--212 (2012)

\bibitem{nah2017deep}
Nah, S., Hyun~Kim, T., Mu~Lee, K.: Deep multi-scale convolutional neural network for dynamic scene deblurring. In: Proceedings of the IEEE Conference on Computer Vision and Pattern Recognition (CVPR) (2017)

\bibitem{nichol2021glide}
Nichol, A., Dhariwal, P., Ramesh, A., Shyam, P., Mishkin, P., McGrew, B., Sutskever, I., Chen, M.: {GLIDE}: Towards photorealistic image generation and editing with text-guided diffusion models (2021)

\bibitem{pan2021exploiting}
Pan, X., Zhan, X., Dai, B., Lin, D., Loy, C.C., Luo, P.: Exploiting deep generative prior for versatile image restoration and manipulation. IEEE Transactions on Pattern Analysis and Machine Intelligence (TPAMI)  \textbf{44}(11),  7474--7489 (2021)

\bibitem{park2023all}
Park, D., Lee, B.H., Chun, S.Y.: {All-in-One} image restoration for unknown degradations using adaptive discriminative filters for specific degradations. In: Proceedings of the IEEE Conference on Computer Vision and Pattern Recognition (CVPR) (2023)

\bibitem{potlapalli2023promptir}
Potlapalli, V., Zamir, S.W., Khan, S., Khan, F.S.: Promptir: Prompting for all-in-one blind image restoration. In: Proceedings of Advances in Neural Information Processing Systems (NeurIPS) (2023)

\bibitem{singletaskqian2018attentive}
Qian, R., Tan, R.T., Yang, W., Su, J., Liu, J.: Attentive generative adversarial network for raindrop removal from a single image. In: Proceedings of the IEEE Conference on Computer Vision and Pattern Recognition (CVPR) (2018)

\bibitem{qian2018attentive}
Qian, R., Tan, R.T., Yang, W., Su, J., Liu, J.: Attentive generative adversarial network for raindrop removal from a single image. In: Proceedings of the IEEE Conference on Computer Vision and Pattern Recognition (CVPR) (2018)

\bibitem{singletaskqin2020ffa}
Qin, X., Wang, Z., Bai, Y., Xie, X., Jia, H.: {FFA-Net}: Feature fusion attention network for single image dehazing. In: Proceedings of the AAAI Conference on Artificial Intelligence (2020)

\bibitem{quan2021removing}
Quan, R., Yu, X., Liang, Y., Yang, Y.: Removing raindrops and rain streaks in one go. In: Proceedings of the IEEE Conference on Computer Vision and Pattern Recognition (CVPR) (2021)

\bibitem{singletaskquan2020self2self}
Quan, Y., Chen, M., Pang, T., Ji, H.: {Self2Self} with dropout: Learning self-supervised denoising from single image. In: Proceedings of the IEEE Conference on Computer Vision and Pattern Recognition (CVPR) (2020)

\bibitem{radford2021learning}
Radford, A., Kim, J.W., Hallacy, C., Ramesh, A., Goh, G., Agarwal, S., Sastry, G., Askell, A., Mishkin, P., Clark, J., et~al.: Learning transferable visual models from natural language supervision. In: Proceedings of International Conference on Machine Learning (ICML) (2021)

\bibitem{ramesh2022hierarchical}
Ramesh, A., Dhariwal, P., Nichol, A., Chu, C., Chen, M.: Hierarchical text-conditional image generation with clip latents. arXiv preprint arXiv:2204.06125  (2022)

\bibitem{singletaskren2019progressive}
Ren, D., Zuo, W., Hu, Q., Zhu, P., Meng, D.: Progressive image deraining networks: A better and simpler baseline. In: Proceedings of the IEEE Conference on Computer Vision and Pattern Recognition (CVPR) (2019)

\bibitem{rombach2022high}
Rombach, R., Blattmann, A., Lorenz, D., Esser, P., Ommer, B.: High-resolution image synthesis with latent diffusion models. In: Proceedings of the IEEE Conference on Computer Vision and Pattern Recognition (CVPR) (2022)

\bibitem{saharia2022photorealistic}
Saharia, C., Chan, W., Saxena, S., Li, L., Whang, J., Denton, E.L., Ghasemipour, K., Gontijo~Lopes, R., Karagol~Ayan, B., Salimans, T., et~al.: Photorealistic text-to-image diffusion models with deep language understanding. Proceedings of Advances in Neural Information Processing Systems (NeurIPS)  (2022)

\bibitem{sheikh2006statistical}
Sheikh, H.R., Sabir, M.F., Bovik, A.C.: A statistical evaluation of recent full reference image quality assessment algorithms. IEEE Transactions on Image Processing (TIP)  (2006)

\bibitem{singh2022flava}
Singh, A., Hu, R., Goswami, V., Couairon, G., Galuba, W., Rohrbach, M., Kiela, D.: Flava: A foundational language and vision alignment model. In: Proceedings of the IEEE Conference on Computer Vision and Pattern Recognition (CVPR) (2022)

\bibitem{singletasksong2023vision}
Song, Y., He, Z., Qian, H., Du, X.: Vision transformers for single image dehazing. IEEE Transactions on Image Processing (TIP)  (2023)

\bibitem{talebi2018nima}
Talebi, H., Milanfar, P.: Nima: Neural image assessment. IEEE Transactions on Image Processing (TIP)  \textbf{27}(8),  3998--4011 (2018)

\bibitem{tao2022df}
Tao, M., Tang, H., Wu, F., Jing, X.Y., Bao, B.K., Xu, C.: {DF-GAN}: A simple and effective baseline for text-to-image synthesis. In: Proceedings of the IEEE Conference on Computer Vision and Pattern Recognition (CVPR) (2022)

\bibitem{tsai2007frank}
Tsai, M.F., Liu, T.Y., Qin, T., Chen, H.H., Ma, W.Y.: {FRank}: a ranking method with fidelity loss. In: ACM SIGGRAPH Conference Proceedings (2007)

\bibitem{valanarasu2022transweather}
Valanarasu, J.M.J., Yasarla, R., Patel, V.M.: Transweather: Transformer-based restoration of images degraded by adverse weather conditions. In: Proceedings of the IEEE Conference on Computer Vision and Pattern Recognition (CVPR) (2022)

\bibitem{van2017neural}
Van Den~Oord, A., Vinyals, O., et~al.: Neural discrete representation learning. Proceedings of Advances in Neural Information Processing Systems (NeurIPS)  (2017)

\bibitem{wang2022exploring}
Wang, J., Chan, K.C., Loy, C.C.: Exploring {CLIP} for assessing the look and feel of images. In: Proceedings of the AAAI Conference on Artificial Intelligence (2023)

\bibitem{singletaskwang2023exploiting}
Wang, J., Yue, Z., Zhou, S., Chan, K.C., Loy, C.C.: Exploiting diffusion prior for real-world image super-resolution. In: arXiv preprint arXiv:2305.07015 (2023)

\bibitem{wang2023stableSR}
Wang, J., Yue, Z., Zhou, S., Chan, K.C., Loy, C.C.: Exploiting diffusion prior for real-world image super-resolution. In: arXiv preprint arXiv:2305.07015 (2023)

\bibitem{wang2022ofa}
Wang, P., Yang, A., Men, R., Lin, J., Bai, S., Li, Z., Ma, J., Zhou, C., Zhou, J., Yang, H.: Ofa: Unifying architectures, tasks, and modalities through a simple sequence-to-sequence learning framework. In: Proceedings of International Conference on Machine Learning (ICML) (2022)

\bibitem{wang2021real}
Wang, X., Xie, L., Dong, C., Shan, Y.: {Real-ESRGAN}: Training real-world blind super-resolution with pure synthetic data. In: Proceedings of the IEEE International Conference on Computer Vision (ICCV) (2021)

\bibitem{wang2018esrgan}
Wang, X., Yu, K., Wu, S., Gu, J., Liu, Y., Dong, C., Qiao, Y., Change~Loy, C.: Esrgan: Enhanced super-resolution generative adversarial networks. In: Proceedings of European Conferences on Computer Vision (ECCV) Workshops (2018)

\bibitem{wang2023ddnm}
Wang, Y., Yu, J., Zhang, J.: Zero-shot image restoration using denoising diffusion null-space model. In: Proceedings of International Conference on Learning Representations (ICLR) (2023)

\bibitem{wei2018deep}
Wei, C., Wang, W., Yang, W., Liu, J.: Deep retinex decomposition for low-light enhancement. Proceedings of The British Machine Vision Conference (BMVC)  (2018)

\bibitem{whang2022deblurring}
Whang, J., Delbracio, M., Talebi, H., Saharia, C., Dimakis, A.G., Milanfar, P.: Deblurring via stochastic refinement. In: Proceedings of the IEEE Conference on Computer Vision and Pattern Recognition (CVPR) (2022)

\bibitem{xu2018real}
Xu, J., Li, H., Liang, Z., Zhang, D., Zhang, L.: Real-world noisy image denoising: A new benchmark. arXiv preprint arXiv:1804.02603  (2018)

\bibitem{xu2018attngan}
Xu, T., Zhang, P., Huang, Q., Zhang, H., Gan, Z., Huang, X., He, X.: {AttnGAN}: Fine-grained text to image generation with attentional generative adversarial networks. In: Proceedings of the IEEE Conference on Computer Vision and Pattern Recognition (CVPR) (2018)

\bibitem{yang2021gan}
Yang, T., Ren, P., Xie, X., Zhang, L.: {GAN} prior embedded network for blind face restoration in the wild. In: Proceedings of the IEEE Conference on Computer Vision and Pattern Recognition (CVPR) (2021)

\bibitem{yang2017deep}
Yang, W., Tan, R.T., Feng, J., Liu, J., Guo, Z., Yan, S.: Deep joint rain detection and removal from a single image. In: Proceedings of the IEEE Conference on Computer Vision and Pattern Recognition (CVPR) (2017)

\bibitem{yao2019attention}
Yao, X., She, D., Zhao, S., Liang, J., Lai, Y.K., Yang, J.: Attention-aware polarity sensitive embedding for affective image retrieval. In: Proceedings of the IEEE Conference on Computer Vision and Pattern Recognition (CVPR) (2019)

\bibitem{ye2021improving}
Ye, H., Yang, X., Takac, M., Sunderraman, R., Ji, S.: Improving text-to-image synthesis using contrastive learning. Proceedings of The British Machine Vision Conference (BMVC)  (2021)

\bibitem{yu2024scaling}
Yu, F., Gu, J., Li, Z., Hu, J., Kong, X., Wang, X., He, J., Qiao, Y., Dong, C.: Scaling up to excellence: Practicing model scaling for photo-realistic image restoration in the wild  (2024)

\bibitem{zamir2022restormer}
Zamir, S.W., Arora, A., Khan, S., Hayat, M., Khan, F.S., Yang, M.H.: Restormer: Efficient transformer for high-resolution image restoration. In: Proceedings of the IEEE Conference on Computer Vision and Pattern Recognition (CVPR) (2022)

\bibitem{zhang2021cross}
Zhang, H., Koh, J.Y., Baldridge, J., Lee, H., Yang, Y.: Cross-modal contrastive learning for text-to-image generation. In: Proceedings of the IEEE Conference on Computer Vision and Pattern Recognition (CVPR) (2021)

\bibitem{zhang2023ingredient}
Zhang, J., Huang, J., Yao, M., Yang, Z., Yu, H., Zhou, M., Zhao, F.: Ingredient-oriented multi-degradation learning for image restoration. In: Proceedings of the IEEE Conference on Computer Vision and Pattern Recognition (CVPR) (2023)

\bibitem{zhang2023brush}
Zhang, L., Chen, X., Wang, Y., Lu, Y., Qiao, Y.: Brush your text: Synthesize any scene text on images via diffusion model. arXiv preprint arXiv:2312.12232  (2023)

\bibitem{zhang2018perceptual}
Zhang, R., Isola, P., Efros, A.A., Shechtman, E., Wang, O.: The unreasonable effectiveness of deep features as a perceptual metric. In: Proceedings of the IEEE Conference on Computer Vision and Pattern Recognition (CVPR) (2018)

\bibitem{8576582}
Zhang, W., Ma, K., Yan, J., Deng, D., Wang, Z.: Blind image quality assessment using a deep bilinear convolutional neural network. IEEE Transactions on Circuits and Systems for Video Technology  \textbf{30}(1),  36--47 (2020)

\bibitem{singletaskzhang2021star}
Zhang, Z., Jiang, Y., Jiang, J., Wang, X., Luo, P., Gu, J.: {STAR}: A structure-aware lightweight transformer for real-time image enhancement. In: Proceedings of the IEEE International Conference on Computer Vision (ICCV) (2021)

\bibitem{singletaskzhang2023real}
Zhang, Z., Jiang, Y., Shao, W., Wang, X., Luo, P., Lin, K., Gu, J.: Real-time controllable denoising for image and video. In: Proceedings of the IEEE Conference on Computer Vision and Pattern Recognition (CVPR) (2023)

\bibitem{zhang2023sine}
Zhang, Z., Han, L., Ghosh, A., Metaxas, D.N., Ren, J.: {SINE}: Single image editing with text-to-image diffusion models. In: Proceedings of the IEEE Conference on Computer Vision and Pattern Recognition (CVPR) (2023)

\bibitem{zhou2022towards}
Zhou, S., Chan, K., Li, C., Loy, C.C.: Towards robust blind face restoration with codebook lookup transformer. Proceedings of Advances in Neural Information Processing Systems (NeurIPS)  (2022)

\bibitem{zhou2022lednet}
Zhou, S., Li, C., Change~Loy, C.: Lednet: Joint low-light enhancement and deblurring in the dark. In: Proceedings of European Conferences on Computer Vision (ECCV) (2022)

\bibitem{zhou2021image}
Zhou, Y., Ren, D., Emerton, N., Lim, S., Large, T.: Image restoration for under-display camera. In: Proceedings of the IEEE Conference on Computer Vision and Pattern Recognition (CVPR) (2021)

\bibitem{zhou2020image}
Zhou, Y., Ren, D., Emerton, N., Lim, S., Large, T.: Image restoration for under-display camera. Proceedings of the IEEE Conference on Computer Vision and Pattern Recognition (CVPR)  (2021)

\bibitem{zhu2019dm}
Zhu, M., Pan, P., Chen, W., Yang, Y.: {DM-GAN}: Dynamic memory generative adversarial networks for text-to-image synthesis. In: Proceedings of the IEEE Conference on Computer Vision and Pattern Recognition (CVPR) (2019)

\bibitem{zhu2022uni}
Zhu, X., Zhu, J., Li, H., Wu, X., Li, H., Wang, X., Dai, J.: Uni-perceiver: Pre-training unified architecture for generic perception for zero-shot and few-shot tasks. In: Proceedings of the IEEE Conference on Computer Vision and Pattern Recognition (CVPR) (2022)

\end{thebibliography}
}
\end{document}